%% file: neurips_2022.tex
\documentclass{article}

\usepackage[preprint,nonatbib]{neurips_2022}

\usepackage[utf8]{inputenc} %
\usepackage[T1]{fontenc}    %
\usepackage{hyperref}       %
\usepackage{url}            %
\usepackage{booktabs}       %
\usepackage{amsfonts}       %
\usepackage{nicefrac}       %
\usepackage{microtype}      %
\usepackage{xcolor}         %
\usepackage{graphicx}  %
\usepackage{paralist}
\usepackage{makecell}
\usepackage{adjustbox}
\usepackage{multirow}
\usepackage{colortbl}
\usepackage{pifont}
\usepackage{amsmath}
\usepackage{floatrow}
\usepackage{stfloats}
\usepackage{xspace}
\usepackage{xcolor}
\usepackage{rotating}
\usepackage{algorithm}
\usepackage[noend]{algpseudocode}
\usepackage{wrapfig}
\usepackage{bbm}
\usepackage{subcaption}
\usepackage{etoc}
\usepackage[font=small]{caption}
\def\etal{\emph{et al}.\xspace}
\definecolor{Gray}{gray}{0.9}
\newcommand{\method}{\texttt{PACMAC}\xspace}
\newcommand{\source}{\mathcal{S}}  %
\newcommand{\target}{\mathcal{T}}  %

\newcommand{\numclasses}{C}  %
\newcommand{\committeesize}{$k$}  %
\newcommand{\maskingratio}{$mr$}  %
\newcommand{\featext}{\phi}  %
\newcommand{\model}{f}  %
\newcommand{\threshold}{$T$}
\newcommand{\modelparams}{\Theta}  %

\DeclareMathOperator*{\argmax}{argmax}

\newfloatcommand{capbtabbox}{table}
\newfloatcommand{capbfigbox}{figure}

\newcommand*{\TODO}{\textcolor{black}}

\title{Adapting Self-Supervised Vision Transformers by Probing Attention-Conditioned Masking Consistency}

\author{
    \textbf{Viraj Prabhu}$^{*}$ \qquad
    \textbf{Sriram Yenamandra}\thanks{Equal contribution. Correspondence to Viraj Prabhu \texttt{<virajp@gatech.edu>}.} \qquad
    \textbf{Aaditya Singh} \qquad 
    \textbf{Judy Hoffman} \qquad \\
    \small{Georgia Institute of Technology \quad}
}

\begin{document}

\maketitle

\vspace{-10pt}
\begin{abstract}
  Visual domain adaptation (DA) seeks to transfer trained models to unseen, unlabeled domains across distribution shift, but approaches typically focus on adapting convolutional neural network architectures initialized with supervised ImageNet representations. In this work, we shift focus to adapting modern architectures for object recognition -- the increasingly popular Vision Transformer (ViT) -- and modern pretraining based on self-supervised learning (SSL). Inspired by the design of recent SSL approaches based on learning from partial image inputs generated via masking or cropping -- either by learning to predict the missing pixels, or learning representational invariances to such augmentations -- we propose \method, a simple two-stage adaptation algorithm for self-supervised ViTs. \method first performs in-domain SSL on pooled source and target data to learn task-discriminative features, and then probes the model's predictive consistency across a set of partial target inputs generated via a novel attention-conditioned masking strategy, to identify reliable candidates for self-training. Our simple approach leads to consistent performance gains over competing methods that use ViTs and self-supervised initializations on standard object recognition benchmarks. Our code is available at \url{https://github.com/virajprabhu/PACMAC}.
\end{abstract}
\vspace{-10pt}

\input{sections/intro}

\input{sections/relwork}

\input{sections/approach}

\input{sections/experiments}

\input{sections/conclusion}

{\small
\bibliographystyle{ieeetr}
\bibliography{main}
}

\input{sections/appendix}

\end{document}

%% file: sections/intro.tex
\vspace{-5pt}
\section{Introduction}
\label{sec:intro}
\vspace{-5pt}

    Deep models struggle to generalize to visual domains that deviate from the ones on which they were trained~\cite{torralba2011unbiased}. Unsupervised domain adaptation~\cite{saenko2010adapting,tzeng2014deep,long2015learning,tzeng2017adversarial,ganin2015unsupervised,long2018conditional,hoffman2018cycada,prabhu2021sentry} focuses on adapting models trained on labeled source domains to unseen and unlabeled target domains, 
    but most approaches still focus on adapting convolutional neural network (CNN) architectures initialized with supervised  representations, typically from ImageNet~\cite{russakovsky2015imagenet}. In this work, we seek to ``modernize'' domain adaptation by focusing on adapting state-of-the-art architectures for object recognition -- Vision Transformers (ViTs)~\cite{dosovitskiy2020image} -- initialized with modern pretraining strategies based on self-supervised learning (SSL).
    
    Vision Transformer (ViT)~\cite{dosovitskiy2020image} architectures have recently gained traction as an effective alternative to CNNs for computer vision tasks~\cite{khan2021transformers}, achieving impressive performance despite fewer inductive biases. 
    With their in-built self-attention mechanisms, and improved calibration under distribution shift over their CNN counterparts~\cite{minderer2021revisiting}, ViTs may be particularly well-suited to domain adaptation~\cite{yang2021tvt,xu2021cdtrans}. 

    Similarly, self-supervised representation learning is rapidly replacing supervised learning as the de-facto pretraining strategy for deep networks, due to its improved scalability (unlabeled data is easier to collect) and generality (domain-specific SSL is often a preferable alternative to one-fits-all ImageNet pretraining~\cite{raghu2019transfusion,azizi2021big}). Many successful methods that optimize proxy objectives based on pretext tasks~\cite{doersch2015unsupervised,wang2015unsupervised,noroozi2016unsupervised}, instance discrimination~\cite{wu2018unsupervised,chen2020simple,he2020momentum,caron2020unsupervised}, self-distillation~\cite{caron2021emerging}, or image reconstruction~\cite{he2021masked}, have been proposed.     
    
    Despite the increasing popularity of ViTs and SSL, to our knowledge no prior work has focused on adapting self-supervised ViTs.
    Recent work has proposed additional in-domain contrastive learning on the pooled source and target domain as a stronger initialization for DA methods~\cite{kim2021cds,shen2022connect}, but focus on adapting CNNs. We follow these works to perform additional in-domain pretraining and find that it leads to learning task-discriminative features with ViTs as well. \TODO{Importantly, Shen~\emph{et al.} find that such contrastive pretraining learns features that disentangle domain and class information and can generalize to the target \emph{without} being domain invariant, subverting traditional wisdom in UDA. We experimentally find that this leads to domain-adversarial methods for UDA consistently underperforming when applied to ViTs initialized with SSL representations, suggesting the need for specialized solutions. Our work attempts to fill this gap.}
    
    Concurrent work has also studied the problem of adapting ViTs~\cite{yang2021tvt,xu2021cdtrans}, but focuses on supervised rather than self-supervised initializations. Crucially, none of these prior works propose an adaptation strategy specifically designed for adapting self-supervised initializations, whereas we
    explicitly incorporate components of the SSL pretraining in our adaptation algorithm for better transfer.  

    Concretely, we note that several recent state-of-the-art SSL methods for ViTs focus on learning from partial inputs~\cite{chen2020simple,he2020momentum,he2021masked,caron2020unsupervised,caron2021emerging} generated via masking or cropping strategies. In the context of vision transformers, such methods learn to either reconstruct the missing information~\cite{he2021masked,assran2022masked} or to be invariant to such cropping~\cite{caron2021emerging}. We find that the \emph{predictive consistency} of a model initialized with such representations across partial inputs generated via masking acts as an effective self-supervised reliability measure. Further, rather than masking inputs randomly, we leverage the ViT attention mechanism to perform \emph{attention-conditioned masking}, and generate a set of disjoint masks corresponding to highly-attended regions of the input image. We then probe the reliability of a given target instance by measuring the model's predictive consistency across such masked images and the original image, and mark target instances with high consistency as reliable. We call our selection strategy Probing Attention-Conditioned Masking Consistency (\method).

    We further mark instances with high confidence as reliable, and then selectively self-train the model against the current prediction on target instances identified as reliable via either scheme. This leads to an easy to implement selective self-training adaptation algorithm that outperforms competing methods on standard benchmarks. 
    We make the following contributions:
    \begin{itemize}
        \item We propose an algorithm to adapt self-supervised ViTs to unseen domains that i) performs in-domain SSL on the pooled source and target data, and ii) self-trains the model on target instances identified to be reliable by probing the model's predictive consistency across a set of attention-conditioned masks applied to the target image, in combination with confidence.
        \item Our approach leads to consistent performance improvements over prior DA methods that use ViTs and self-supervised initializations on the OfficeHome~\cite{venkateswara2017deep}, DomainNet~\cite{peng2019moment}, and VisDA~\cite{peng2017visda} adaptation benchmarks for object recognition.
    \end{itemize}

%% file: sections/relwork.tex
\vspace{-5pt}
\section{Related Work}
\label{sec:relwork}
\vspace{-5pt}

\noindent\textbf{Unsupervised domain adaptation (UDA).} UDA seeks to transfer a model trained on a labeled source domain to an unlabeled target domain across distribution shift. A variety of successful approaches based on aligning feature spaces via optimizing domain divergence measures~\cite{long2015learning,tzeng2014deep} and distribution matching via domain adversarial learning~\cite{ganin2015unsupervised,tzeng2017adversarial,long2018conditional} (often incorporating additional pixel-level matching constraints), have been proposed. Recently, selective self-training~\cite{tan2020class,prabhu2021sentry} on the model's predictions on target data has emerged as a simple and effective UDA technique. 

With the upsurge in adoption of Vision Transformer (ViT~\cite{dosovitskiy2020image}) architectures for object recognition, some recent works specifically focus on adapting ViTs~\cite{yang2021tvt,xu2021cdtrans}. These methods leverage the ViT attention mechanism for incorporating patch-level transferability~\cite{yang2021tvt} and category-level feature alignment~\cite{xu2021cdtrans}. However, these works focus on adapting models initialized with supervised ImageNet weights. In this work, we propose an adaptation algorithm designed for self-supervised ViTs.

\noindent\textbf{Self-supervised learning.}
Self-supervised learning (SSL) leverages unlabeled data to learn representations that can be transferred efficiently to downstream tasks. Several approaches based on pretext tasks~\cite{doersch2015unsupervised,wang2015unsupervised,noroozi2016unsupervised}, masked-image modeling~\cite{he2021masked, bao2021beit}, context prediction~\cite{pathak2016context} and instance discrimination~\cite{chen2020simple, he2020momentum}, have been proposed to learn strong semantic representations in the absence of labelled data. Some of the state-of-the-art SSL methods are designed for ViTs: DINO~\cite{caron2021emerging} performs self-distillation by training a student network that is given a strongly augmented image as input to match the output of a teacher network that sees a global view of the image. MAE~\cite{he2021masked} is a masked image modeling approach that learns to predict a complete image given only a fraction of the image patches as input. 
A concurrent work, MSN~\cite{assran2022masked} improves upon DINO by passing in a masked version of an augmented image to the student network. We derive inspiration from the local-to-global self-supervisory signal used by these methods in designing our adaptation algorithm, but instead of encouraging reconstruction or invariance to missing information, \emph{measure} the model's predictive confidence under targeted missing information as a self-supervised probe to determine reliability.

\noindent\textbf{Self-supervised learning for domain adaptation.} While most UDA methods focus on adapting models initialized with supervised initializations, typically from ImageNet, a few recent works have studied self-supervised domain adaptation~\cite{assran2022masked,kim2021cds}. Kim~\emph{et al.}~\cite{kim2021cds} propose CDS, a pretraining strategy for domain adaptation that makes use of in-domain contrastive learning in conjunction with cross-domain matching as a superior initialization alternative to ImageNet pretraining. Shen~\emph{et al.}~\cite{shen2022connect} study the transferability of representations learned via additional in-domain contrastive learning on the source and target domain, finding that such pretrained features can be transferred effectively to the target by simply finetuning on the source, despite not being domain invariant.
However, these methods restrict their experiments to ResNet~\cite{he2016deep} architectures while we work focus on adapting ViTs. Crucially, neither of these works propose a DA algorithm catered to the model's self-supervised initialization, whereas we do so explicitly.

%% file: sections/approach.tex
\vspace{-10pt}
\section{Approach}
\label{sec:approach}
\vspace{-5pt}

\subsection{Notation}
\vspace{-5pt}

\noindent
Let $\mathcal{X}$ and $\mathcal{Y}$ denote input and output spaces. In unsupervised domain adaptation (UDA) we are given access to labeled source instances $(\mathbf{x}_\source, y_\source) \sim \mathcal{P}_\source(\mathcal{X}, \mathcal{Y})$, and unlabeled target instances $\mathbf{x}_\target \sim \mathcal{P}_\target(\mathcal{X})$, where $\source$ and $\target$ correspond to source and target domains. The goal is to learn a model $\model=h(\featext(.))$ ($\featext$ denotes the encoder and $h$ denotes classifier), parameterized by $\modelparams$ with minimum error on the target dataset. In our experiments, $\modelparams$ is parameterized as a Vision Transformer or ViT~\cite{dosovitskiy2020image}), which takes as input a linear embedding of of N image patches in addition to a class token embedding. These N+1 embeddings are then appended with positional encodings and passed through several transformer layers, each of which comprises of a sequence of multi-headed self-attention (with M attention heads), multilayer perceptron, and layernorm modules. Features from the final encoder layer are typically fed to a classifier layer and used to predict the output. We consider UDA in the context of $C$-way image classification: the inputs $\mathbf{x}$ are images, and labels $y$ are categorical variables $y \in \{1, 2, .. , \numclasses \}$. 
For an image $\mathbf{x}$, let $p_{\modelparams}(y|\mathbf{x})$ denote the final probabilistic output from the model. 
For a target instance $\mathbf{x}_\target \sim \mathcal{P}_\target(\mathcal{X})$, we compute a ``pseudolabel'' as ${\hat{y}} = \argmax p_{\modelparams}(y | \mathbf{x}_\target)$. 

\vspace{-5pt}
\subsection{Preliminaries}
\vspace{-5pt}

\noindent\textbf{Self-supervised Learning for Vision Transformers.} We discuss SSL methods for ViTs that learn from partial inputs, under two popular formulations: i) Masked Image Modeling, wherein the model is given a partial image and trained to predict the missing content (either at a pixel or token level) via learning a masked or denoising autoencoder~\cite{he2021masked,xie2021simmim,wei2021masked,bao2021beit}. ii) Joint-embedding networks, which seek to learn a model that produces similar features for different views of a given image, across strong cropping and additional augmentation~\cite{caron2021emerging,assran2022masked}. Let $m(\mathbf{x}_\target)$ denote the partial image generated from target image $\mathbf{x}_\target$ under transformation $m(.)$ which could correspond to either a masking or augmentation strategy. Broadly speaking, SSL corresponds to minimizing an objective of the form $\mathcal{L}_{SSL}(\featext(m(\mathbf{x}_\target)), \featext(\mathbf{x}_\target))$, where $\mathcal{L}_{SSL}$ could correspond to a reconstruction or invariance based objective operating on encoded features.

\noindent\textbf{Self-training for Unsupervised DA.} Self-training for UDA typically involves training the model on its predictions on unlabeled target data (``pseudolabels''), optimizing either a conditional entropy minimization~\cite{grandvalet2004semi} or  cross-entropy objective. However, under a domain shift many of the model's predictions may initially be incorrect~\cite{chen2019progressive,jiang2020implicit,prabhu2021sentry} (especially under severe distribution shift), and unconstrained self-training may lead to amplifying model errors.

To combat this, a few recent works have proposed selective self-training on predictions that have a high-likelihood of being correct. Recent works propose to identify such reliable predictions either based on model confidence~\cite{tan2020class,zou2018unsupervised}, predictive consistency under data augmentation~\cite{prabhu2021sentry,kim2022ev}, or combinations of the two~\cite{prabhu2021s4t}. Let $r(\mathbf{x}_\target)$ denote a binary reliability value for target instance ${x}_\target$. We optimize the following selective self-training objective on target data $L_{SST}$:

\begin{equation}
    \mathcal{L}_{SST} = \mathbb{E}_{(\mathbf{x}_\target, \hat{y}_\target) \sim \mathcal{P}_\target} [r(\mathbf{x}_\target) \mathcal{L}_{CE} (\model(\mathbf{x}_\target), \hat{y}_\target)]    
    \label{eq:ce}
\end{equation}    

In this work, we propose an alternative selection criterion designed for self-supervised representations based on predictive consistency under attention-conditioned masking, which we now introduce.

\begin{figure*}[t]
    \centering
    \includegraphics[width=0.95\linewidth]{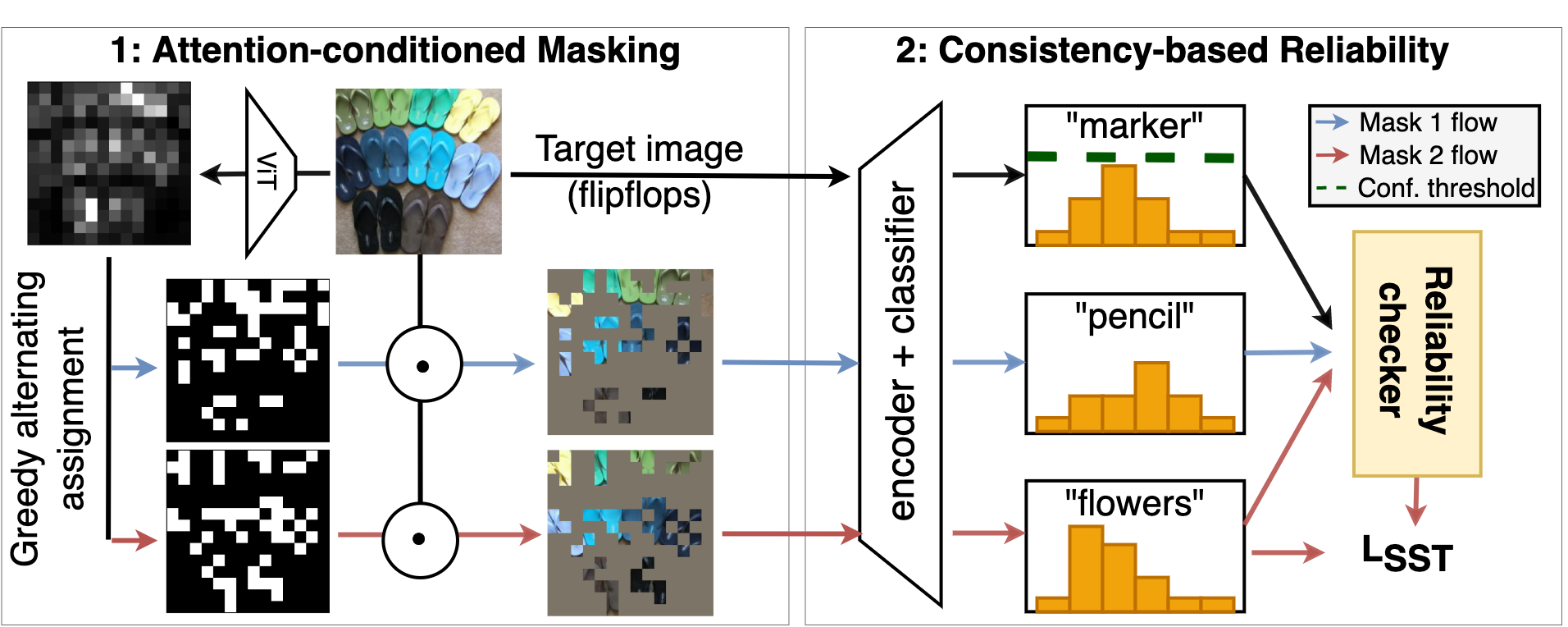}
    \caption{\textbf{Overview of \method.} \textbf{Left.} First, the model's attention over a given target image is used to generate a set of disjoint masks corresponding to highly attended image regions via a greedy assignment strategy, which are then applied to the original image. \textbf{Right.} Next, the model's predictive consistency across the original and masked images is used as a self-supervised reliability measure to select target instances for self-training. }
    \label{fig:approach}
 \end{figure*}

 \vspace{-5pt}
\subsection{\method: Probing Attention-conditioned Masking Consistency for UDA}
\vspace{-5pt}

Following prior work~\cite{shen2022connect,kim2021cds}, we first learn task-discriminative features by by optimizing the following in-domain self-supervised pretraining objective $L_{IDP}$ on pooled source and target data:

\begin{equation}
    \mathcal{L}_{IDP} = \mathbb{E}_{\mathbf{x}_\source \sim \mathcal{P}_\source} [\mathcal{L}_{SSL}(\featext(m(\mathbf{x}_\source)), \featext(\mathbf{x}_\source))] + \mathbb{E}_{\mathbf{x}_\target \sim \mathcal{P}_\target} [\mathcal{L}_{SSL}(\featext(m(\mathbf{x}_\target)), \featext(\mathbf{x}_\target))]  
    \label{eq:idp}
\end{equation}  

Next, we perform selective self-training. For a target instance $\mathbf{x}_\target \sim \mathcal{P}_{\target}$, we generate a committee of \committeesize masked versions. However, applying random masks may lead to capturing irrelevant background features; instead, we propose \emph{attention-conditioned masking} (see Algo.~\ref{algo:attentionSeededMasking}): we first obtain the model's per-patch attention score $\hat{a}_\target$ over the original image. In practice, we obtain the patchwise self-attention of the class token with respect to all image patches for each attention head, and average across heads. We then sort patches in descending order of attention scores and perform a greedy round-robin assignment to the \committeesize masks, until the desired masking ratio is satisfied. We then apply these masks to the original image to generate masked images $\{m_1(\mathbf{x}_\target), m_2(\mathbf{x}_\target), ..., m_k(\mathbf{x}_\target)\}$. 

We make predictions for each of these $k$ masked versions, and measure \emph{consistency} between the model's prediction for the original image and for each of the $k$ masked versions. If the model's prediction for all masked versions matches its prediction on the original image, we consider the instance as ``reliable'', and as ``unreliable'' if not. Further, we also consider instances as reliable if the model's predictive confidence on the original image is higher than a threshold \threshold. Formally, we assign reliability $r(\mathbf{x}_\target)$ as:
\begin{equation}
	\scalebox{0.9}{$r(\mathbf{x}_\target) = 
	\begin{cases}
		1 & \text{if } \overbrace{\argmax p_{\Theta}(y | m_i(x_\target)) == \argmax p_{\Theta}(y | (x_\target)) \forall i=1 .. k}^{\text{consistent}} \; 
		\text{or} \; 
		\overbrace{\max\; p_{\modelparams}(y | \mathbf{x}_\target) > T}^{\text{confident}} \\
      0  & \text{otherwise} \\
	\end{cases}$}
\end{equation}

Algorithm~\ref{algo:amcon} lists the steps involved in computing the \method objective. Without loss of generality, we minimize the cross-entropy between the model's prediction on the last consistent masked image $m_k(\mathbf{x}_\target)$ and its consistent pseudolabel. We find this to act as an effective data augmentation strategy and provide a stronger learning signal than when using the model's prediction on the original unmasked image. Following prior work we regularize training by optimizing an additional cross-entropy loss over labeled source examples. For loss weight $\alpha$, the full $L_\method$ objective is:

\begin{equation}
    \mathcal{L}_{\method} = \mathbb{E}_{(\mathbf{x}_\source, y_\source) \sim \mathcal{P}_\source} [\mathcal{L}_{CE} (\model(\mathbf{x}_\source), y_\source)] + \alpha \mathbb{E}_{(\mathbf{x}_\target, \hat{y}_\target) \sim \mathcal{P}_\target}  [r(\mathbf{x}_\target) \mathcal{L}_{CE} (\model(m_k(\mathbf{x}_\target)), \hat{y}_\target)]    
    \label{eq:amcon}
\end{equation}    

\begin{algorithm}
    \caption{Attention-conditioned Masking}
    \label{algo:attentionSeededMasking}
    \begin{algorithmic}[1]
    \State Input: $x_\target,$ per-patch attention $\hat{a}_\target$, masking ratio \maskingratio, committee size \committeesize
    \State L $\gets$ (1-\maskingratio) $\times$ len($\hat{a}_\target$)
    \State $\hat{S}_\target$ $\gets$ argsort($\hat{a}_\target$) \Comment{\textcolor{blue}{Sort patch-wise attention}}
    \For{i $\gets$ 1 to k}
        \State $M_i \gets \text{zeros\_like}(\hat{a})$
        \Comment{\textcolor{blue}{Initialize masks}}
    \EndFor
    \State i $\gets$ 0
    \While{i < L}
        \State i $\gets$ i + 1
        \For{j $\gets$ 1 to k}
            \State p $\gets \hat{S}_\target \text{.pop()}$ \Comment{\textcolor{blue}{Get next-most attended patch index}}
            \State $M_j^p \gets 1$ \Comment{\textcolor{blue}{Greedy round-robin assignment}}
        \EndFor
    \EndWhile
    \For{j $\gets$ 1 to k}
    \State $m_j(x_\target) = M_j \odot x_\target$
    \EndFor
    \State Return $\{m_1(x_\target), \dots, m_k(x_\target) \}$
    \end{algorithmic}
    \end{algorithm}

\begin{algorithm}
    \caption{\method Optimization}
    \label{algo:amcon}
    \begin{algorithmic}[1]
    \State Input: SrcLoader, TgtLoader, model parameters $\Theta$, masking ratio \maskingratio, committee size \committeesize, confidence threshold \threshold, loss weight $\alpha$
    \For{epoch $\gets$ 1 to \texttt{MAX\_EPOCH}}
        \For{$x_\source,y_\source$ in SrcLoader and $x_\target$ in TgtLoader}
            \State $\hat{p}_\target, \hat{y}_\target  \gets \max p_{\Theta}(y | x_\target)$\Comment{\textcolor{blue}{Get top-1 confidence and prediction on clean image}}
            \State $\hat{a}_\target  \gets \text{att}(p_{\Theta}(y | x_\target)) $\Comment{\textcolor{blue}{Get per-patch attention}}
            \State $\{m_1(x_\target), \dots, m_k(x_\target) \} \gets$ AttentionConditionedMasking($x_\target, \hat{a}_\target$, \maskingratio, \committeesize)
            \State {$\text{C} \gets \{m_i(x_\target)| \hat{y}_\target = \argmax p_{\Theta}(y | m_i(x_\target))  \}_{i=1}^k$}
            \State $\mathcal{L}_{SST} = 0$
            \If{\texttt{len}(\text{C}) $=$ \committeesize $\;\text{OR}\; \hat{p}_\target > \mathbf{T}$ } \Comment{\textcolor{blue}{Consistent or Confident}}
                \State $\mathcal{L}_{SST} \gets \mathcal{L}_{CE}(p_{\Theta}(y | \text{C}\text{.last()}), \hat{y}_\target)$
            \EndIf
            \State Minimize $\mathcal{L}_{CE}(x_\source,y_\source) + \alpha \mathcal{L}_{SST} $
            \EndFor
    \EndFor
    \end{algorithmic}
    \end{algorithm}

%% file: sections/experiments.tex
\vspace{-5pt}
\section{Experiments}
\label{sec:experiments}
\vspace{-5pt}

\par\noindent We first describe our datasets and metrics (Sec.~\ref{ref:datasets}), implementation details (Sec.~\ref{ref:implementation}), and baselines (Sec.~\ref{ref:baselines}). Next, we present results (Sec.~\ref{ref:results}), method ablations (Sec.~\ref{ref:ablations}), and analysis (Sec.~\ref{ref:analysis}).

\vspace{-5pt}
\subsection{Datasets and Metrics}
~\label{ref:datasets}
\vspace{-5pt}

We evaluate \method on three classification benchmarks for domain adaptation: i) \noindent\textbf{OfficeHome.} OfficeHome~\cite{venkateswara2017deep} is a classification-based benchmark comprising of 12 shifts spanning 65 categories of objects found in home and office environments. It consists of 4 domains: Real-world (\textbf{Rw}), Clipart (\textbf{Cl}), Product (\textbf{Pr}), and Art (\textbf{Ar}). 
ii) \noindent\textbf{DomainNet.} DomainNet~\cite{peng2019moment} is a large benchmark for adapting object recognition models. Matching prior work~\cite{shen2022connect,prabhu2021sentry}, we use the subset of DomainNet proposed in Tan~\etal~\cite{tan2020class} for our experiments, which reports performance over 12 shifts comprising 40 common classes from 4 domains: Real (\textbf{R}), Clipart (\textbf{C}), Painting (\textbf{P}), and Sketch (\textbf{S}). 
iii) \noindent\textbf{VisDA2017.} The VisDA2017 dataset~\cite{peng2017visda} is large image classification benchmark for synthetic$\rightarrow$real domain adaptation. It consists of over 200k images spanning 12 categories.

\par\noindent\textbf{Metric.} Matching prior work we report standard accuracy on the target test set as our metric.

\vspace{-5pt}
\subsection{Implementation details}
\label{ref:implementation}
\vspace{-5pt}

We use a ViT-base~\cite{dosovitskiy2020image} architecture with 16x16 image patches. We initialize models with official MAE~\cite{he2021masked} and DINO~\cite{caron2021emerging} checkpoints pretrained on ImageNet1K~\cite{russakovsky2015imagenet}, and use their official codebases to perform additional pretraining on the combined source and target domain for 800 epochs (MAE) and 200 epochs (DINO). We linearly scale the learning rate to $4\times10^{-4}$ (MAE) and $5\times10^{-5}$ (DINO) during a 40 epoch warmup phase followed by a cosine decay. We use the AdamW~\cite{loshchilov2018decoupled} optimizer. For \method, we use \committeesize~$=2$, \maskingratio~$=0.75$, \threshold~$=50\%$, and $\alpha=0.1$. We use RandAugment~\cite{cubuk2020randaugment} with $N=3$ and $M=4.0$ during pretraining and $N=1$ and $M=2.0$ during adaptation. On OfficeHome and DomainNet, we perform source finetuning for 100 epochs followed by 100 epochs of DA, and perform 10 epochs of each phase on VisDA. We use a learning rate of $2\times10^{-4}$ and weight decay of $0.05$. We use PyTorch~\cite{paszke2019pytorch} for all experiments.

\begin{table*}[t]
  \begin{center} 
  \setlength{\tabcolsep}{2pt}
  \resizebox{\textwidth}{!}{
  \begin{tabular}{c l c c c c c c c c c c c c c}
      \toprule
      {\bf IN1K Init. } & {\bf Method}  & $\mathbf{A} \rightarrow \mathbf{C}$ & $\mathbf{A} \rightarrow \mathbf{P}$ & $\mathbf{A} \rightarrow \mathbf{R}$ & $\mathbf{C} \rightarrow \mathbf{A}$ & $\mathbf{C} \rightarrow \mathbf{P}$ & $\mathbf{C} \rightarrow \mathbf{R}$ & $\mathbf{P} \rightarrow \mathbf{A}$ & $\mathbf{P} \rightarrow \mathbf{C}$ & $\mathbf{P} \rightarrow \mathbf{R}$ & $\mathbf{R} \rightarrow \mathbf{A}$ & $\mathbf{R} \rightarrow \mathbf{C}$ & $\mathbf{R} \rightarrow \mathbf{P}$ & {\bf AVG} \\
      \midrule
      \multirow{5}{*}{\centering \makecell{MAE~\cite{he2021masked}}} & source  & 46.4	&57.6&	71.0&	51.1&	60.0&	62.6&	51.4&	46.9&	70.5&	66.3&	52.2&	77.2&	59.4\\      
      & CDAN~\cite{long2018conditional}  & 45.3	& 58.8	& 69.1& 	51.6& 	60.7& 	61.5& 	53.4& 	45.5& 	72.4& 	67.7& 	49.9& 	78.0& 	59.5 \\        
      & MCC~\cite{jin2020minimum} & 43.9 &	61.2	&70.7&	52.8&	59.9&	62.8&	51.1&	40.3&	70.9&	66.2&	48.3&	76.3&	58.7 \\        
      & SENTRY~\cite{prabhu2021sentry}  & 54.8	&  65.6	& 74.4 & 	56.5	& 65.8& 69.8		& 57.6& 54.9			&  75.5 & 68.9 & 60.0 & 81.6	& 65.5		 \\      
      \rowcolor{Gray}       
      & \method (Ours)  & 58.9	& 68.2	& 74.1& 	60.6& 	67.1& 	67.2& 	57.3& 	59.2& 	74.4& 	68.6& 	63.9& 	82.7& 	\textbf{66.8} \\
      \midrule
      \multirow{6}{*}{\centering \makecell{DINO~\cite{caron2021emerging}}}
      & source  & 53.1	& 65.0& 	75.2& 	62.0& 	66.2& 	70.4& 	60.8& 	50.5& 	77.0& 	72.8& 	53.9& 	81.2& 	65.7\\
      & CDAN~\cite{long2018conditional}  & 36.9	& 59.9&	71.8&	44.0&	61.3&	63.8&	51.1 &	36.3 &	76.6 &	69.6 &	45.5&	81.2&	58.2\\
      & MCC~\cite{jin2020minimum} & 44.4& 	74.2& 	79.6& 	61.9& 	67.6& 	72.4& 	63.0& 	40.1& 	79.2& 	73.3& 	47.1& 	82.8& 	65.5 \\        
      & TVT~\cite{yang2021tvt}  & 48.3	& 65.7& 	73.6& 	60.6& 	68.8& 	64.6& 	57.1& 	44.1& 	75.4& 	71.0& 	53.7& 	77.2& 	63.3\\
      & SENTRY~\cite{prabhu2021sentry}  & 59.5	& 72.0	& 76.8& 	66.1& 	71.1& 	73.4& 	63.7& 	56.2& 	77.8& 	72.4& 	63.0& 	81.9& 	69.5 \\      
      \rowcolor{Gray}
      & \method (Ours)  & 54.9 &	74.7	&79.3&	65.7 &	74.0&	74.5&	63.3&	55.8&	79.2&	73.1&	58.4&	83.9&	\textbf{69.7} \\     
      \bottomrule
      \end{tabular}
      }
      \vspace{-7pt}
      \caption{Target test set accuracy on OfficeHome across MAE~\cite{he2021masked} and DINO~\cite{caron2021emerging} pretraining.}\label{tab:officehome}
      \vspace{-7pt}
  \end{center}
\end{table*}

\vspace{-10pt}
\subsection{Baselines and SSL strategies}
~\label{ref:baselines}
\vspace{-5pt}

\noindent\textbf{DA baselines:} In the absence of DA baselines designed for adapting self-supervised ViTs, we compare against diverse strategies based on domain adversarial learning (CDAN~\cite{long2018conditional}), minimizing classifier-confusion (MCC~\cite{jin2020minimum}), and self-training (SENTRY~\cite{prabhu2021s4t}), originally proposed for CNNs with supervised initializations. We also benchmark a concurrent method for adapting ViTs (TVT~\cite{yang2021tvt}) with supervised initializations.
\noindent\textbf{1) CDAN~\cite{long2018conditional}:} CDAN improves upon domain adversarial learning with multilinear conditioning, by capturing cross-covariance between feature representations and classifier predictions to improve discriminability. Recent work~\cite{kim2022broad} finds CDAN to outperform more recent DA methods when combined with new architectures like ViTs. 
\noindent\textbf{2) MCC~\cite{jin2020minimum}:} Minimum Classifier Confusion is a non-adversarial DA method that aligns domains by minimizing pairwise class confusion on the target domain estimated from model predictions.
\noindent\textbf{3) SENTRY~\cite{prabhu2021sentry}}: SENTRY measures model predictive consistency across randomly augmented versions of each target image and selectively minimizes entropy to increase predictive confidence on highly consistent instances, while maximizing it to decrease confidence on highly inconsistent ones. \noindent\textbf{4) TVT~\cite{yang2021tvt}}: Transferable Vision Transformer (TVT) injects a learned transferability measure into the transformer attention blocks, in addition to performing global domain-adversarial alignment and discriminative clustering.

\noindent\textbf{SSL Strategies:} As discussed in Sec.\TODO{~\ref{sec:approach}}, \method is primarily designed as an adaptation strategy for self-supervised representations. We report benchmark performance on top of two popular SSL strategies for ViTs (see appx. for more details): \noindent\textbf{1) MAE~\cite{he2021masked}}: Masked Autoencoders (MAE) are a visual self-supervised learning strategy that learns a visual transformer autoencoder to reconstruct images given only a random subset of patches from the original image. By masking out large portions of images ($\sim$75\%), the MAE encoder is shown to learn strong representations that can be effectively finetuned to downstream tasks. \noindent\textbf{2) DINO~\cite{caron2021emerging}}: Self-Distillation with No Labels is a visual self-supervised learning strategy that passes two transformed versions of each image to a student and teacher network respectively, and trains the student network to predict the output of the teacher.

\TODO{In the appendix we also benchmark \method on the OfficeHome benchmark on top of a supervised ImageNet initialization, and observe gains despite not being designed for such initializations.}

\vspace{-5pt}
\subsection{Results}
~\label{ref:results}
\vspace{-5pt}

Tables\ref{tab:officehome}, \ref{tab:domainnet}, and ~\ref{tab:visda} present results for our method on the OfficeHome, DomainNet, and VisDA benchmarks. We observe:

\noindent\textbf{$\triangleright$ Several existing DA methods underperform on top of SSL representations.} On both OfficeHome and DomainNet, we observe existing DA methods (CDAN~\cite{long2018conditional}, MCC~\cite{jin2020minimum}, TVT~\cite{yang2021tvt}) to frequently underperform even the source model on average, despite careful hyperparameter tuning. The fact that many existing DA methods struggle with such initializations suggests the need for specialized solutions, particularly as SSL initializations become more common. SENTRY~\cite{prabhu2021s4t} and \method however offer consistent improvements. In Sec.~\ref{ref:analysis} we further investigate the failure of CDAN on top of SSL representations.

\begin{table*}[t]
  \begin{center} 
  \setlength{\tabcolsep}{2pt}
  \resizebox{\textwidth}{!}{
  \begin{tabular}{c l c c c c c c c c c c c c c}
      \toprule
      {\bf IN1K Init. } & {\bf Method}  & $\mathbf{R} \rightarrow \mathbf{C}$ & $\mathbf{R} \rightarrow \mathbf{P}$ & $\mathbf{R} \rightarrow \mathbf{S}$ & $\mathbf{C} \rightarrow \mathbf{R}$ & $\mathbf{C} \rightarrow \mathbf{P}$ & $\mathbf{C} \rightarrow \mathbf{S}$ & $\mathbf{P} \rightarrow \mathbf{R}$ & $\mathbf{P} \rightarrow \mathbf{C}$ & $\mathbf{P} \rightarrow \mathbf{S}$ & $\mathbf{S} \rightarrow \mathbf{R}$ & $\mathbf{S} \rightarrow \mathbf{C}$ & $\mathbf{S} \rightarrow \mathbf{P}$ & {\bf AVG} \\
      \midrule
      \multirow{4}{*}{\centering \makecell{MAE~\cite{he2021masked}}} & source & 71.0&	77.6&	62.9&	73.7&	61.5&	63.3&	82.4&	63.1&	66.1&	76.6&	71.9&	69.6&	70.1 \\
      &CDAN~\cite{long2018conditional} & 72.2	& 74.5& 	59.3& 	80.6& 	57.3& 	59.2& 	78.5& 	57.4& 	61.2& 	81.4& 	73.2& 	69.4& 	67.7 \\
      &SENTRY~\cite{prabhu2021sentry} & 84.2	& 82.8& 	76.4& 	86.9& 	77.1& 	74.1& 	86.9& 	76.2& 	73.3& 	88.8& 	81.6& 	77.6& 	80.5\\ 
      \rowcolor{Gray}
      &\method (Ours)  & 86.0	& 81.9& 	78.8& 	86.0& 	74.8& 	76.3& 	87.4& 	84.0& 	77.5& 	85.2& 	83.1& 	78.3& 	\textbf{81.6} \\        
      \midrule
      \multirow{5}{*}{\centering \makecell{DINO~\cite{caron2021emerging}}} & source & 75.7&	82.8&	68.2&	81.9&	73.9&	71.1&	82.6&	70.6&	69.5&	80.8&	76.9&	78.1&	76.0\\
      &CDAN~\cite{long2018conditional} & 66.1 &	72.6	&50.8&	75.6	&45.2&	41.1&	79.1&	48.5&	43.4&	77.1&	61.9&	58.9&	58.2 \\
      &TVT~\cite{yang2021tvt} & 70.4& 	79.7& 	64.2& 	80.2& 	68.1& 	65.2& 	81.6& 	61.9& 	65.6& 	80.3& 	71.4& 	74.1& 	71.9\\        
      &SENTRY~\cite{prabhu2021sentry} & 81.8 &	80.9&	73.4&	89.1& 79.4	&	75.8&	86.5	& 75.6&	71.6&	88.3&	81.9& 82.4	&  80.4\\          
      \rowcolor{Gray}        
      &\method (Ours) & 80.7& 	82.9& 	82.0& 	85.7& 	78.8& 	78.3& 	87.3& 	75.5& 	75.2& 	84.7& 	79.6& 	82.0& 	\textbf{81.0} \\        
      \bottomrule
      \end{tabular}
      }
      \vspace{-7pt}
      \caption{Target test set accuracy on DomainNet across MAE~\cite{he2021masked} and DINO~\cite{caron2021emerging} pretraining.}\label{tab:domainnet}
      \vspace{-7pt}
  \end{center}
\end{table*}

\newlength{\oldintextsep}
\setlength{\oldintextsep}{\intextsep}

\setlength\intextsep{0pt}
\begin{wraptable}{t}{4cm}
  \vspace{-5pt}
  \caption{Target accuracy on VisDA, with a DINO~\cite{caron2021emerging} init.}\label{tab:visda}
  \begin{tabular}{l c}
    \toprule
    {\bf Method} & {\bf Acc.} \\
    \midrule
    source & 68.9 \\
    CDAN~\cite{long2018conditional} & 72.4 \\
    TVT~\cite{yang2021tvt} & 65.9 \\  
    SENTRY~\cite{prabhu2021sentry} & 76.0 \\
    \rowcolor{Gray}
    \method (Ours) & \textbf{81.0} \\
    \bottomrule
    \end{tabular}
  \end{wraptable} 
  
\noindent\textbf{$\triangleright$ \method outperforms competing methods across benchmarks and initializations.} Averaged over 12 shifts on OfficeHome, \method improves over the next-best method by 1.3\% (MAE) and 0.2\% (DINO). On DomainNet, we observe 12-shift average gains of 1.1\% (MAE) and 0.6\% (DINO). On VisDA, we observe a gain of 5\%. We note that though gains over the next best method (SENTRY~\cite{prabhu2021sentry}) are relatively small across some settings, our method is \emph{substantially} simpler, and does not make use of diversity regularizers, class-balancing on the source and target, or an entropy maximization loss.

\subsection{Ablating \method}
~\label{ref:ablations}

\begin{table}[!ht]    
  \centering
  \RawFloats
  \begin{subfloatrow} 
  \ffigbox[\FBwidth][][!htbp]
{
  \resizebox{.4\textwidth}{!}{
      \setlength{\tabcolsep}{3pt}
      \begin{tabular}{l c c c c}
        \toprule
        \multirow{2}{*}{\centering \bf PT domains} & \multicolumn{2}{c}{\centering \bf MAE} & \multicolumn{2}{c}{\centering \bf DINO} \\
        \cmidrule(l{4pt}r{4pt}){2-3}
        \cmidrule(l{4pt}r{4pt}){4-5}
        &  kNN & FT & kNN & FT \\
        \midrule
       IN1K PT & 29.6 & 60.0 & 57.9 & 66.2\\
       + S PT & 32.7 & 55.6 & - & - \\
       + T PT & 39.3 & 64.2 & - & - \\
       + S+T PT & 41.4 & 66.2 & 56.5 & 68.9 \\
       \;\;+ dom. decoders & 19.7 &57.4  & N/A & N/A \\
        \bottomrule
        \end{tabular}
}}
  {
  \caption{\small Pretraining ablations}
  \label{tab:ablate_pretrain}
  }
  \hspace{-0.75cm}
  \ffigbox[\FBwidth][][!htbp]
{
  \resizebox{.25\textwidth}{!}{
      \setlength{\tabcolsep}{3pt}
      \begin{tabular}{l c c }
        \toprule        
         &  \textbf{MAE} & \textbf{DINO} \\
        \midrule
        source & 60.0 & 66.2 \\
        \rowcolor{Gray}
       \method & 67.1 & 74.0 \\
       \; - S+T PT & 61.2  & 72.2\\      
        \bottomrule
        \end{tabular}
}}
  {
  \caption{\small Vary initialization}
  \label{tab:ablate_init}
  }
  \hspace{-0.5cm}
  \ffigbox[\FBwidth][][!htbp]    
  {
    \vspace{-10pt}
  \resizebox{.32\textwidth}{!}{
      \setlength{\tabcolsep}{3pt}
      \begin{tabular}{l c}
        \toprule
        \textbf{select on target} & \textbf{Acc.} \\
        \midrule
       all & 68.6 \\
       confident & 71.3 \\
       consistent & 73.7 \\
       consistent AND confident & 72.2 \\
       \rowcolor{Gray}
       consistent OR confident & 74.0 \\
       \midrule
       correct (oracle) & 94.0 \\
        \bottomrule
        \end{tabular}      
}
}
{
    \caption{\small Ablating selection strategy}
    \label{tab:ablate_sel}
  }
  \end{subfloatrow}
  \caption{\small OfficeHome Cl$\to$Pr. Gray is ours: a) \textbf{Ablating pretraining.} Cross-domain kNN and finetuning (FT) accuracies for MAE and DINO strategies (S=Source, T=Target, PT=Pretrain, dom.=domain). b) \textbf{Varying \method init.} Transfer accuracy with and without S+T pretraining. c) Ablating selection strategy with DINO init.}
  \label{tab:ablate_pret_sel}
\end{table}

In Tables~\ref{tab:ablate_pret_sel}-~\ref{tab:ablate_con} we ablate \method with a DINO initialization on OfficeHome Cl$\to$Pr. We observe:

\noindent\textbf{$\triangleright$ In-domain pretraining helps.(\ref{tab:ablate_pretrain})} We report two metrics: cross-domain k-Nearest Neighbor (k=7) accuracy on the target domain using source domain embeddings from the trained encoder (kNN column), and target accuracy after finetuning on the source (FT column). Across MAE and DINO initializations, we observe additional in-domain pretraining on the pooled source and target domain to improve finetuning performance (Row 4 v/s 1, +6.2\% FT acc. with MAE, +2.7\% with DINO). With MAE, we additionally try pretraining only on source (S, Row 2) or target (T, Row 3) domains and find them to underperform S+T pretraining. We also try S+T pretraining with MAE by learning separate decoders to reconstruct source and target images, and find that to underperform no pretraining (Row 5 v/s 1, -2.6\%). Interestingly, we find pretraining to improve kNN accuracy in all cases except Row 5 with MAE, indicating better domain alignment in encoder feature space. With DINO, we observe higher kNN accuracies than MAE, but see a slight drop after in-domain pretraining.

\noindent\textbf{$\triangleright$ \method benefits from S+T pretraining (\ref{tab:ablate_init})}
We apply \method to ImageNet SSL features \emph{without} S+T pretraining and observe worse accuracy (Row 3 v/s 2, -5.9\% with MAE, -1.8\% with DINO).

\noindent\textbf{$\triangleright$ Combining masking consistency and confidence is an effective selection measure.} In Table~\ref{tab:ablate_sel}, we first self-train on all (Row 1) or only confident (model confidence $>$ 50\%, Row 2) target examples. We find both to underperform selection based on our proposed attention-conditioned masking consistency scheme (Row 3, \textbf{+5.1\%} and \textbf{+2.4\%}). We further combine our consistency measure with confidence and find selecting instances marked as reliable by atleast one measure to perform best (Row 5). In Row 6, as we report upper-bound performance of an oracle method which only self-trains on correct pseudolabels, and find it to achieve a high accuracy of 94\%. We also try self-training on the original rather than masked image (underperforms by 3.2\%).

\noindent\textbf{$\triangleright$ \method's gains are not simply due to a stronger initialization.} To verify this, we swap out \method's attention-conditioned masking consistency scheme with the committee consistency across augmentation scheme used by the next best performing method, SENTRY~\cite{prabhu2021sentry}. We match original hyperparameters and use a committee size of 3, RandAugment~\cite{cubuk2020randaugment} parameters of N=3 and M=2.0, and majority voting. We find this obtains an 71.8\% v/s 74.0\% (\textbf{+2.2\%}) with \method. This indicates the increased efficacy of our selection strategy over SENTRY's for adapting self-supervised ViTs.

\noindent\textbf{$\triangleright$ Attention-conditioning helps, as does committee consistency.} In Table~\ref{tab:ablate_com_att}, we vary the committee size \committeesize\; and masking strategy (random v/s attention-conditioned masking). As seen, attention conditioning \emph{consistently} improves upon random masking across \committeesize\; (\textbf{+0.5 to 1.8\%}).

\noindent\textbf{$\triangleright$ Ablating masking ratio and confidence threshold.} In Tables~\ref{tab:ablate_mr} and ~\ref{tab:ablate_conf}, we vary the masking ratio \maskingratio and confidence threshold \threshold, and find \maskingratio=75\% and \threshold=50\% to work best.

\begin{table}[!t]    
  \centering
  \vspace{5pt}
  \RawFloats
  \begin{subfloatrow} 
  \ffigbox[\FBwidth][][!htbp]    
  {
  \resizebox{.32\textwidth}{!}{
      \setlength{\tabcolsep}{4pt}
      \begin{tabular}{c | c c c}
          \toprule

          & $k$=1& $k$=2 & $k$=3 (U/M)\\
          \hline
         rnd. mask & 68.1 & 72.6 & 71.3/72.0 \\
         att. mask & 68.6 & \cellcolor{Gray}74.0 & 73.1/72.8 \\
          \bottomrule
          \end{tabular}        
}
}
{
    \caption{\small Ablating consistency checker}
    \label{tab:ablate_com_att}
  }
\ffigbox[\FBwidth][][!htbp]
{
  \resizebox{.24\textwidth}{!}{
      \setlength{\tabcolsep}{4pt}
      \begin{tabular}{c c}
        \toprule
        \textbf{Masking ratio} & \textbf{Acc.} \\
        \midrule
        50\% & 72.3 \\          
        \rowcolor{Gray}
        75\% & 74.0 \\          
        90\% & 71.0 \\              
        \bottomrule
        \end{tabular}
}}
  {
  \caption{\small Vary masking ratio}
  \label{tab:ablate_mr}
  }
\hspace{0.2cm}
  \ffigbox[\FBwidth][][b]
{
  \resizebox{.24\textwidth}{!}{
      \setlength{\tabcolsep}{4pt}
      \begin{tabular}{c c}
          \toprule
          \textbf{conf. thresh.}  & \textbf{Acc.} \\
          \midrule
          25\% & 73.7 \\
          \rowcolor{Gray}
          50\% & 74.0 \\
          75\% & 72.9 \\
          \bottomrule
          \end{tabular}
}}
  {
  \caption{\small Vary confidence threshold}
  \label{tab:ablate_conf}
  }
  \end{subfloatrow}
  \vspace{-7pt}
  \caption{\small \textbf{Ablating the consistency checker on OH Cl$\rightarrow$Pr}: \textbf{a)} Varying committee size ($k$), masking strategy (rnd.=random, att.=attention, U=unanimous, M=majority voting). \textbf{b)} Varying masking ratio \textbf{c)} Varying confidence threshold for selection. Gray is our method.}
  \vspace*{-20pt}
  \label{tab:ablate_con}
\end{table}

\begin{figure}[h]
  \centering  
    \begin{subfigure}[t]{0.48\textwidth}
      \centering
      \includegraphics[width=\linewidth]{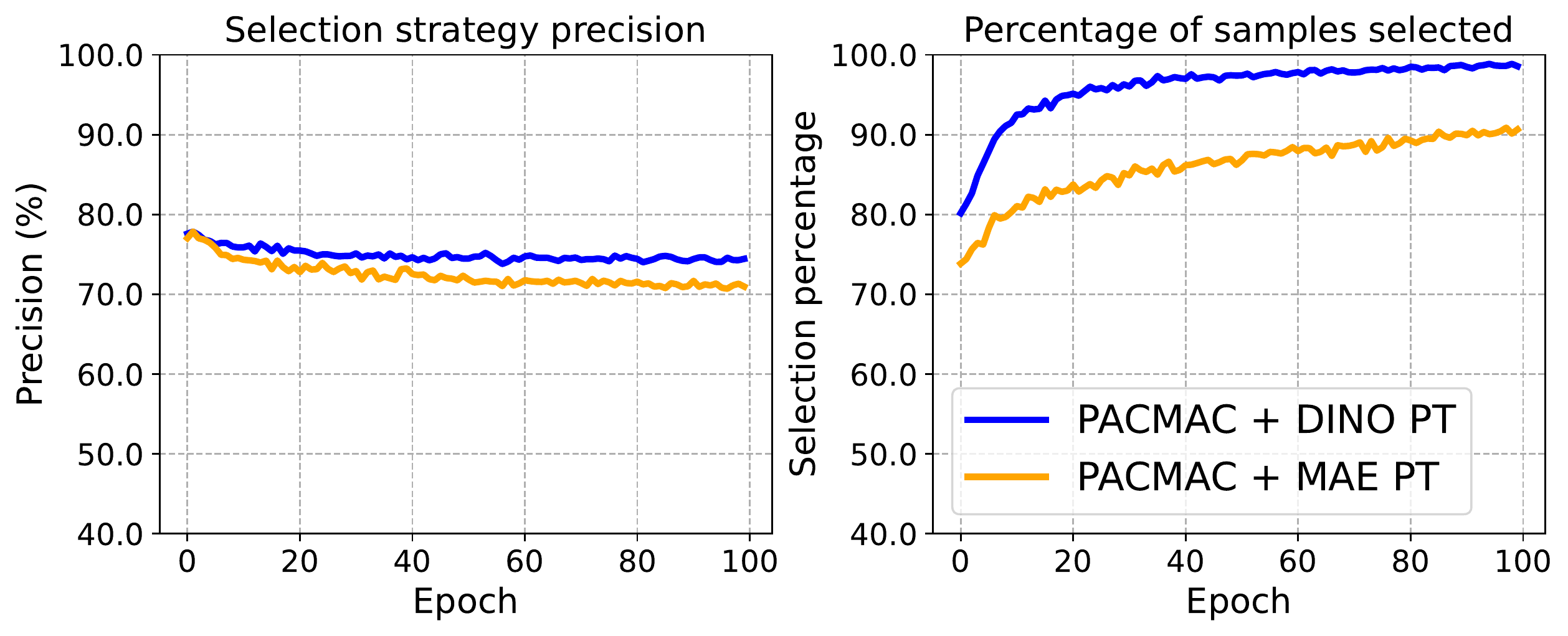}
      \caption[]%
      {{\small \textbf{Left.} Precision of our selection strategy with DINO (blue) and MAE (yellow) initializations. \textbf{Right.} Percentage of target examples selected for self-training across epochs. }}  
      \label{fig:breakdown}
      \end{subfigure}
    \begin{subfigure}[t]{0.5\textwidth}
      \vspace*{-2.5cm}
    \resizebox{\textwidth}{!}{
      \setlength{\tabcolsep}{2pt}
      \begin{tabular}{l c c c c c}
        \toprule
        PT  & err(diff. D) & err(same C,  & err(diff. C, & err(diff. C & DA \\
        strat. & (\%) &  same D)(\%)  & same D)(\%) & diff. D)(\%) & score \\
        \midrule
       MAE & 8.1&	10.6	&6.6	&3.3 & 3.8 \\
       DINO & 10.4 &	8.5	& 2.6 &	1.0 & 5.9 \\
       \midrule
       Sup. & 14.2	 & 11.9 &	3.2 &	1.6 & 8.5 \\
        \bottomrule
        \end{tabular}
    }    
    \vspace{10pt}
    \caption[]%
    {{\small Error of linear classifier trained to distinguish features for: \textbf{C2.} Domains. \textbf{C3.} Same class, different domains. \textbf{C3.} Different class, same domains. \textbf{C4.} Different class, different domains. \textbf{C5.} Domain alignment score. Results are averaged across all OfficeHome shifts. }}  
    \label{tab:cdan}
  \end{subfigure}
    \label{fig:analysis}
    \caption{\textbf{Left}. Evaluating reliability estimation. \textbf{Right}. Understanding SSL initialization.}
  \vspace{-7pt}
\end{figure}

\begin{figure*}[t]
  \centering
  \includegraphics[width=1.0\linewidth]{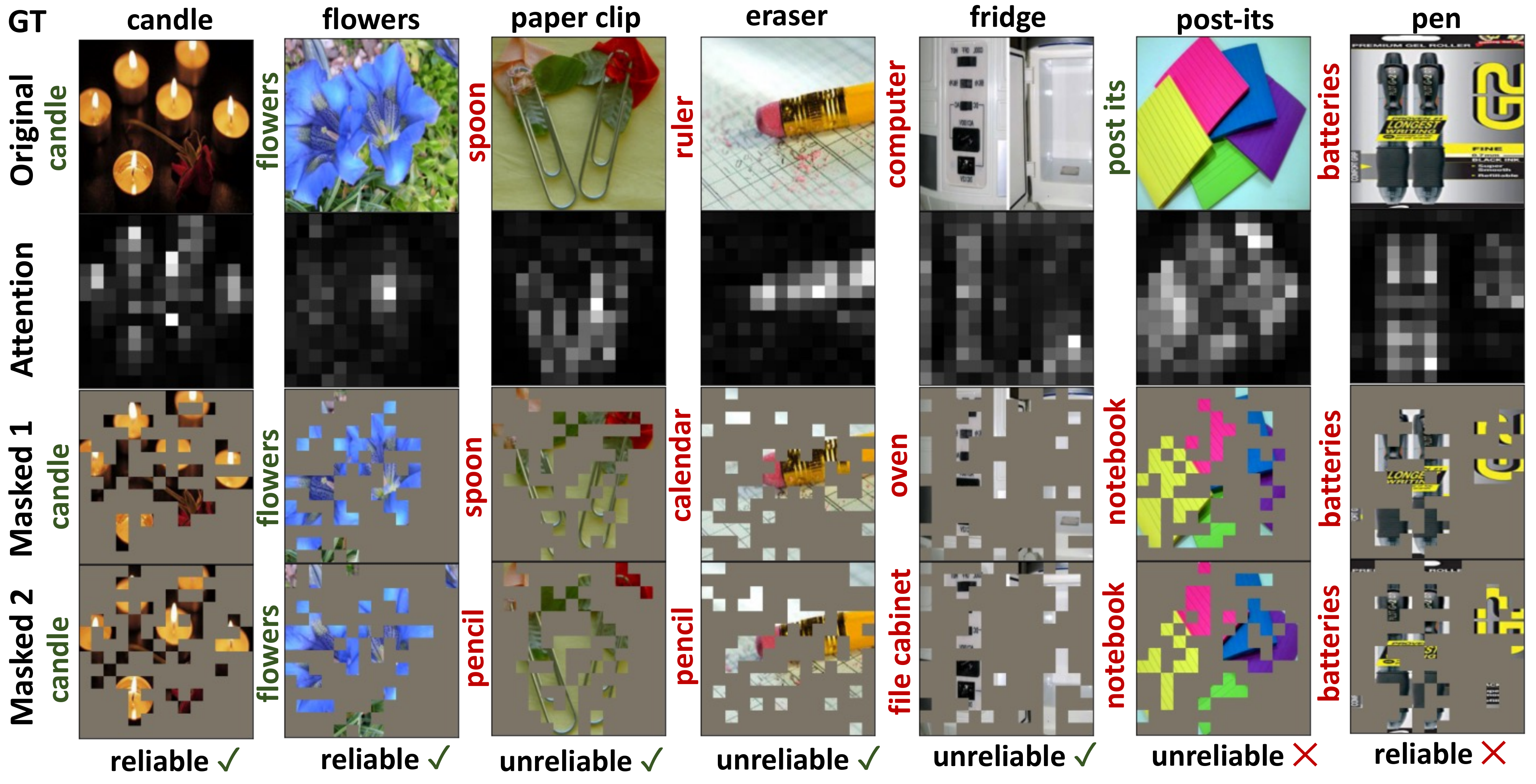}
  \vspace{-5pt}
  \caption{\textbf{Visualizing \method.} Row 1: Ground truth label. Row 2: Original image. Row 3: Per-patch attention. Rows 5-6: Masked images. We include model predictions to the left of each image, color coded as green (correct) and red (incorrect). Row 6: \method consistency (tick and cross denote correct and incorrect assessment).}
  \vspace{-15pt}
  \label{fig:qualitative}
\end{figure*}

\vspace{-10pt}
\subsection{Analyzing \method}
~\label{ref:analysis}
\vspace{-20pt}

\textbf{Evaluating reliability estimation.} In Fig.~\ref{fig:breakdown}, we evaluate our proposed reliability estimation scheme across MAE and DINO initializations. We observe a high precision (70-80\% across both) across epochs, indicating that our method correctly identifies reliable instances with a low false-positive rate (per-class analysis in appx.). We also find that the percentage of target instances selected selected for self-training increases over time, and is particularly high with DINO pretraining.

\TODO{\textbf{Understanding SSL initialization: Why does CDAN~\cite{long2018conditional} fail?} In Tab.~\ref{tab:cdan}, we contrast self-supervised (MAE and DINO) and supervised ImageNet initialization, by reporting the error of linear classifiers (averaged over all OfficeHome shifts) trained to distinguish class token features from: \textbf{C2.} Different domains. We observe considerably higher error for the supervised init. indicating better domain alignment. \textbf{C3.} Same class but different domains (low error with DINO indicates less per-class domain alignment) \textbf{C4.} different classes but same domain (high error with MAE indicates more inter-class confusion), and \textbf{C5.} different domains and classes. \textbf{C6.} We define a domain alignment (DA) score = C3-max(C4, C5) -- higher means that cross-domain examples from different domain but same class are on average closer than examples from different classes and either the same or different domains. Intuitively, a high DA score will translate to better performance for CDAN\footnote{To validate this, we measure Pearson correlation between DA score and transfer accuracy with CDAN on all OH shifts, and observe high correlation: 0.9 (MAE), 0.86 (DINO), and \TODO{TBD} (sup.).}, which seeks to make logits-conditioned source and target features indistinguishable. We find that self-supervised inits indeed have a low DA score, possibly explaining why CDAN applied to such features fails.}

\textbf{Visualizing attention-conditioned masking consistency.} In Fig.~\ref{fig:qualitative}, we visualize our proposed attention-conditioned masking consistency scheme for random target images from OfficeHome. As seen, attention-conditioning ensures that the masks correspond to disjoint, highly attended regions of the target image (Row 2). 
Columns 1-5 illustrate examples for which our selection strategy is correctly able to identify reliable (Cols 1-2) and unreliable instances (Cols 3-5). Cols 6-7 denote failure cases, with the first showing a false negative (correctly classified instance being misidentified as unreliable) and the second denoting a false positive.

\textbf{Effect of in-domain pretraining on per-class kNN accuracy.} In the appendix, we report per-category cross-domain kNN accuracy on the OfficeHome Cl$\to$Pr shift before and after in-domain pretraining across MAE and DINO initializations. We find that accuracy improves on several classes in both cases, particularly so for MAE. While prior work has observed benefits from pretraining for supervised CNNs~\cite{shen2022connect,kim2021cds}, we corroborate this finding in the context of self-supervised ViTs.

In the appendix, we analyze the effect of in-domain pretraining on out-of-distribution confidence calibration (observing inconsistent trends across shifts)
, and include t-SNE~\cite{van2008visualizing} visualizations of the feature space learned by encoders before and after in-domain pretraining.

%% file: sections/conclusion.tex
\vspace{-10pt}
\section{Limitations}
\label{sec:limitations}
\vspace{-5pt}

Despite its simplicity, \method requires i) additional in-domain pretraining and ii) additional forward passes over masked target images (but no additional backpropagation), which increases its computational overhead. Further, \method's effectiveness at identifying reliable instances varies across categories (see appx.). In our paper we only experiment with ViTs, and the generality of our approach across architectures is not established. \TODO{Finally, our approach still lags behind methods that use supervised initializations, suggesting the need to rethink SSL with cross-domain transfer in mind}.

\section{Conclusion}
\label{sec:conclusion}

We present \method, a simple algorithm to adapt ViTs initialized with self-supervised representations. \method first performs in-domain pretraining on pooled source and target data, and then uses predictive consistency across partial versions of target images generated via an attention-conditioned masking strategy to judge reliability for selective self-training. \method obtains consistent performance gains over competing methods on standard recognition benchmarks.

\noindent\textbf{Acknowledgements.} This work was supported in part by funding from the DARPA LwLL project and ARL.

%% file: sections/appendix.tex
\appendix

\section{Improving \method performance}

Recall that in Section 4.4 we pointed out that our method \method outperfoms \texttt{SENTRY}~\cite{prabhu2021sentry} without additional diversity regularizers or entropy maximization losses. We now attempt to add these pieces to \method: specifically, we replace the target cross-entropy objective on reliable instances with an entropy minimization loss $\mathcal{L}_{entmin}(\mathbf{x}_\target) =  \sum_{c=1}^{\numclasses} - p_{\modelparams}(y\!=\!c|\mathbf{x}_\target) \log p_{\modelparams}(y\!=\!c|\mathbf{x}_\target)$, optimize an additional information entropy loss to encourage diverse predictions across all target instances $L_{div} = \sum_{c=1}^{\numclasses} p_{\modelparams}(y\!=\!c|\mathbf{x}_\target) \log q(\hat{y}\!=\!c)$ ($q(\hat{y})$ denotes a running average of model predictions, loss weight$=5\times 10^{-4}$), and perform additional entropy maximization to reduce model confidence on unreliable target instances $\mathcal{L}_{entmax}(\mathbf{x}_\target) =  \sum_{c=1}^{\numclasses} + p_{\modelparams}(y\!=\!c|\mathbf{x}_\target) \log p_{\modelparams}(y\!=\!c|\mathbf{x}_\target)$ (loss weight$=1.0$). We denote this method as \method*. 

As shown in Table~\ref{tab:officehome} below, across both MAE~\cite{he2016deep} and DINO~\cite{caron2020unsupervised} initializations this further improves performance by 0.5\% (MAE) and 0.9\%  (DINO) on average.

\begin{table*}[t]
    \begin{center} 
    \setlength{\tabcolsep}{2pt}
    \resizebox{\textwidth}{!}{
    \begin{tabular}{c l c c c c c c c c c c c c c}
        \toprule
        {\bf IN1K Init. } & {\bf Method}  & $\mathbf{A} \rightarrow \mathbf{C}$ & $\mathbf{A} \rightarrow \mathbf{P}$ & $\mathbf{A} \rightarrow \mathbf{R}$ & $\mathbf{C} \rightarrow \mathbf{A}$ & $\mathbf{C} \rightarrow \mathbf{P}$ & $\mathbf{C} \rightarrow \mathbf{R}$ & $\mathbf{P} \rightarrow \mathbf{A}$ & $\mathbf{P} \rightarrow \mathbf{C}$ & $\mathbf{P} \rightarrow \mathbf{R}$ & $\mathbf{R} \rightarrow \mathbf{A}$ & $\mathbf{R} \rightarrow \mathbf{C}$ & $\mathbf{R} \rightarrow \mathbf{P}$ & {\bf AVG} \\
        \midrule
        \multirow{3}{*}{\centering \makecell{MAE~\cite{he2021masked}}} & source  & 46.4	&57.6&	71.0&	51.1&	60.0&	62.6&	51.4&	46.9&	70.5&	66.3&	52.2&	77.2&	59.4\\      
    
        & SENTRY~\cite{prabhu2021sentry}  & 54.8	&  65.6	& 74.4 & 	56.5	& 65.8& 69.8		& 57.6& 54.9			&  75.5 & 68.9 & 60.0 & 81.6	& 65.5		 \\      
        \rowcolor{Gray}       
        & \method (Ours)  & 58.9	& 68.2	& 74.1& 	60.6& 	67.1& 	67.2& 	57.3& 	59.2& 	74.4& 	68.6& 	63.9& 	82.7& 	66.8 \\
        \rowcolor{Gray}
        & \method (Ours)*  & 59.5 &	68.1  &	74.3 &	60.2 &	68.2&	70.1&	57.6&	59.0&	74.5&	67.9&	65.8&	82.4&	\textbf{67.3} \\
        \midrule
        \multirow{3}{*}{\centering \makecell{DINO~\cite{caron2021emerging}}}
        & source  & 53.1	& 65.0& 	75.2& 	62.0& 	66.2& 	70.4& 	60.8& 	50.5& 	77.0& 	72.8& 	53.9& 	81.2& 	65.7\\
        & SENTRY~\cite{prabhu2021sentry}  & 59.5	& 72.0	& 76.8& 	66.1& 	71.1& 	73.4& 	63.7& 	56.2& 	77.8& 	72.4& 	63.0& 	81.9& 	69.5 \\      
        \rowcolor{Gray}
        & \method (Ours)  & 54.9 &	74.7	&79.3&	65.7 &	74.0&	74.5&	63.3&	55.8&	79.2&	73.1&	58.4&	83.9&	69.7 \\     
        \rowcolor{Gray}
        & \method* (Ours)  & 56.6 &	75.2&	79.2&	65.8&	73.3&	74.8&	65.8&	56.8&	79.3&	73.6&	61.9&	85.0&\textbf{70.6} \\     
        \bottomrule
        \end{tabular}
        }
        \vspace{-7pt}
        \caption{\textbf{Improving \method with \texttt{SENTRY} regularizers (denotes as \method*).} Target test set accuracy on OfficeHome across MAE~\cite{he2021masked} and DINO~\cite{caron2021emerging} pretraining.}\label{tab:officehome}
        \vspace{-7pt}
    \end{center}
  \end{table*}

\section{\method: Additional analysis} 

\subsection{Per-class accuracy change}

\begin{figure}[t]
    \centering  
      \begin{subfigure}[t]{0.9\textwidth}
        \centering
        \includegraphics[width=\linewidth]{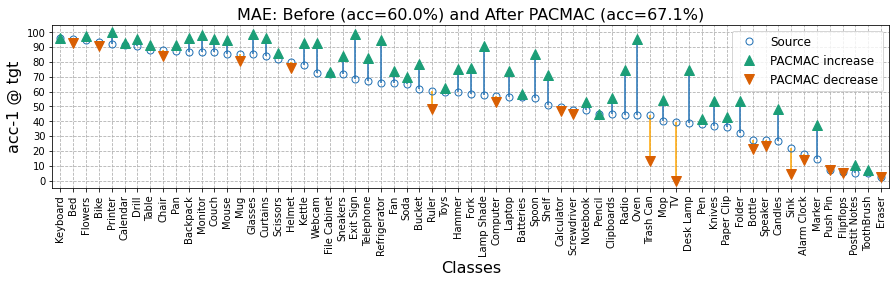}
        \caption[]%
        {{\small \textbf{MAE~\cite{he2016deep}  }}}  
        \label{fig:perclassAccMAE}
        \end{subfigure}
        \begin{subfigure}[t]{0.9\textwidth}
            \centering
        \includegraphics[width=\linewidth]{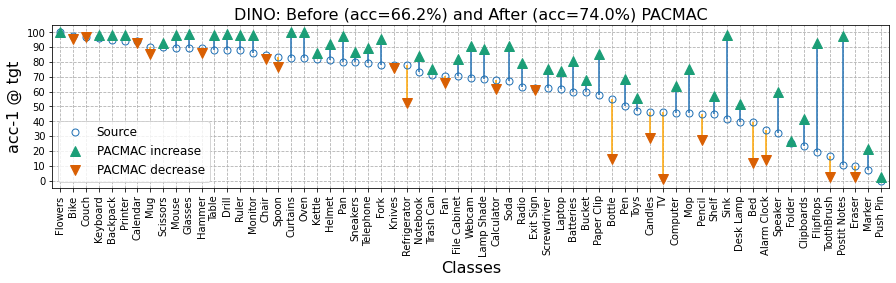}
            \caption[]%
            {{\small \textbf{DINO~\cite{caron2021emerging}}}  
            \label{fig:perclassAccDINO}}
            \end{subfigure} 
      \caption{\textbf{Per-class accuracy with \method}: Target accuracy before and after applying \method on the OfficeHome Clipart$\to$Product shift.}
      \label{fig:perclassAcc}
    \vspace{-7pt}
  \end{figure}

In Fig.~\ref{fig:perclassAcc} we present per-class accuracy changes after applying \method to the source model across MAE and DINO initializations on the OfficeHome Clipart$\to$Product shift. As seen, across both plots \method maintains or improves accuracy across most categories. However, performance for a few categories falls, which we analyze in the next experiment.

\subsection{Reliability checker: Per-class analysis}

\begin{figure*}[h]
    \centering
    \includegraphics[width=.9\linewidth]{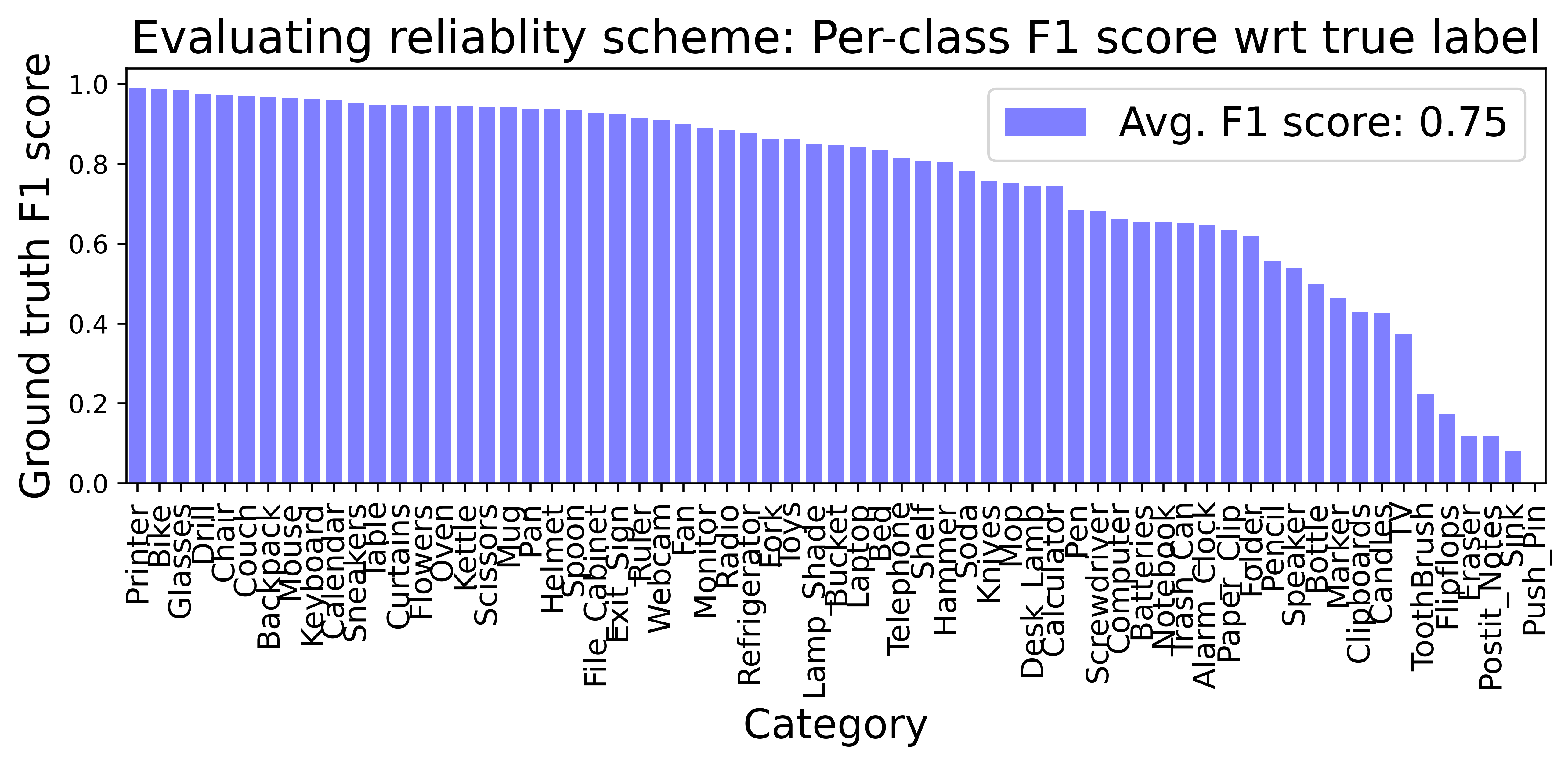}
    \vspace{-5pt}
    \caption{\textbf{Evaluating reliability-checker: Per-class analysis} }
    \label{fig:pca}
  \end{figure*}

In Fig.~\ref{fig:pca} we evaluate the performance of our consistency or confidence based reliability determination scheme on a per-class level. We use a model pretrained on the OfficeHome Clipart$\to$Product shift with DINO, and finetuned on the source domain. We then compute per-class F1 score of the estimated reliability on the target domain so as to capture both precision (how often is a reliable instance actually correct?) and recall (what fraction of correct instances are identified by our method?). As seen, F1 scores are high for a majority of classes. However, performance is noticeably worse on some classes (such as TV, bottle, and alarm clock). Unsurprisingly, we find that model accuracy on these categories also drops after applying \method (Fig.~\ref{fig:perclassAccDINO}).

\subsection{In-domain Pretraining: Per-class accuracy} 

\begin{figure}[t]
    \centering  
      \begin{subfigure}[t]{0.9\textwidth}
        \centering
        \includegraphics[width=\linewidth]{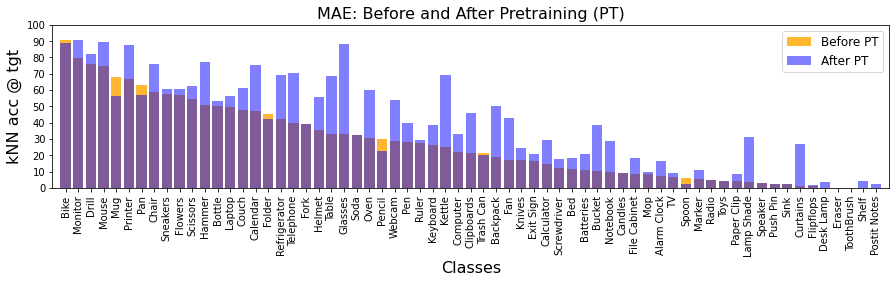}
        \caption[]%
        {{\small \textbf{MAE~\cite{he2016deep}  }}}  
        \label{fig:mae_ood}
        \end{subfigure}
        \begin{subfigure}[t]{0.9\textwidth}
            \centering
            \includegraphics[width=\linewidth]{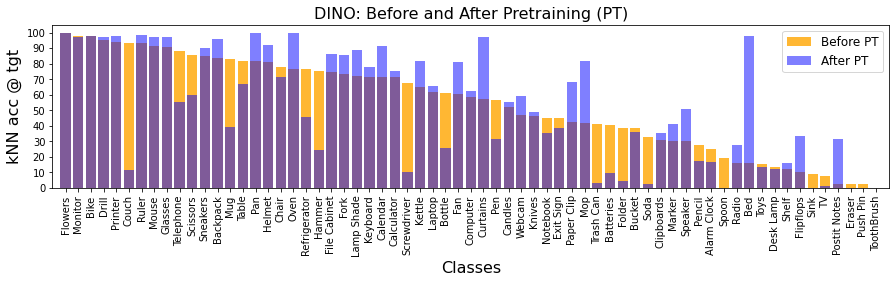}
            \caption[]%
            {{\small \textbf{DINO~\cite{caron2021emerging}}}  
            \label{fig:dino_ood}}
            \end{subfigure} 
      \caption{\textbf{Per-class accuracy}: Cross-domain kNN accuracies after additional in-domain pretraining on the source and target domains}
      \label{fig:perclasskNN}
    \vspace{-7pt}
  \end{figure}

In Fig.~\ref{fig:perclasskNN} we report per-category cross-domain kNN accuracy on the OfficeHome Cl$\to$Pr shift before and after in-domain pretraining across MAE and DINO initializations. We find that accuracy improves on several classes in both cases, particularly so for MAE.

\subsection{In-domain Pretraining: OOD calibration} 

\begin{figure}[b]
    \centering  
      \begin{subfigure}[t]{0.48\textwidth}
        \centering
        \includegraphics[width=\linewidth]{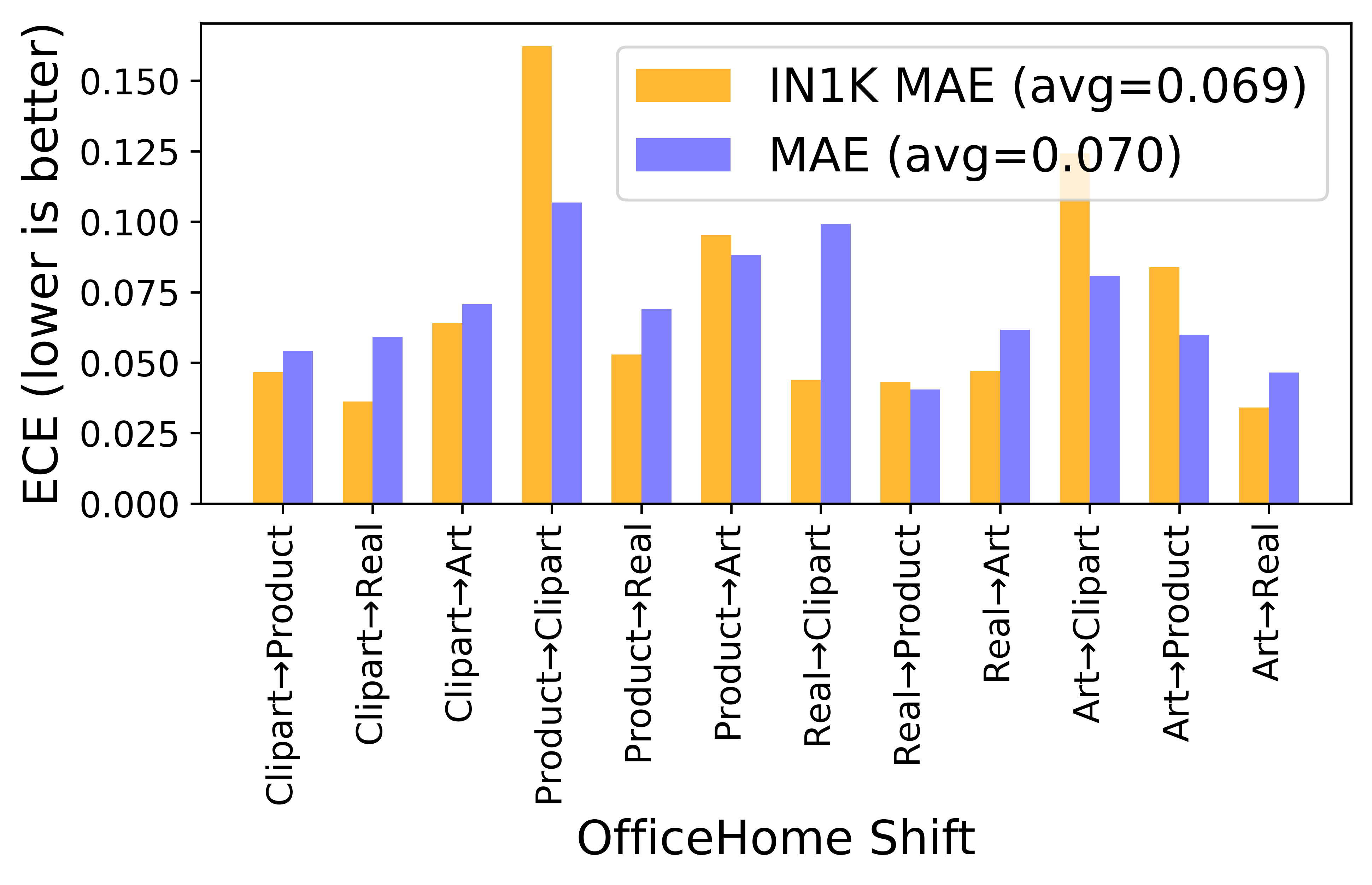}
        \caption[]%
        {{\small \textbf{MAE S+T pretraining}  }}  
        \label{fig:mae_ood}
        \end{subfigure}
        \begin{subfigure}[t]{0.48\textwidth}
            \centering
            \includegraphics[width=\linewidth]{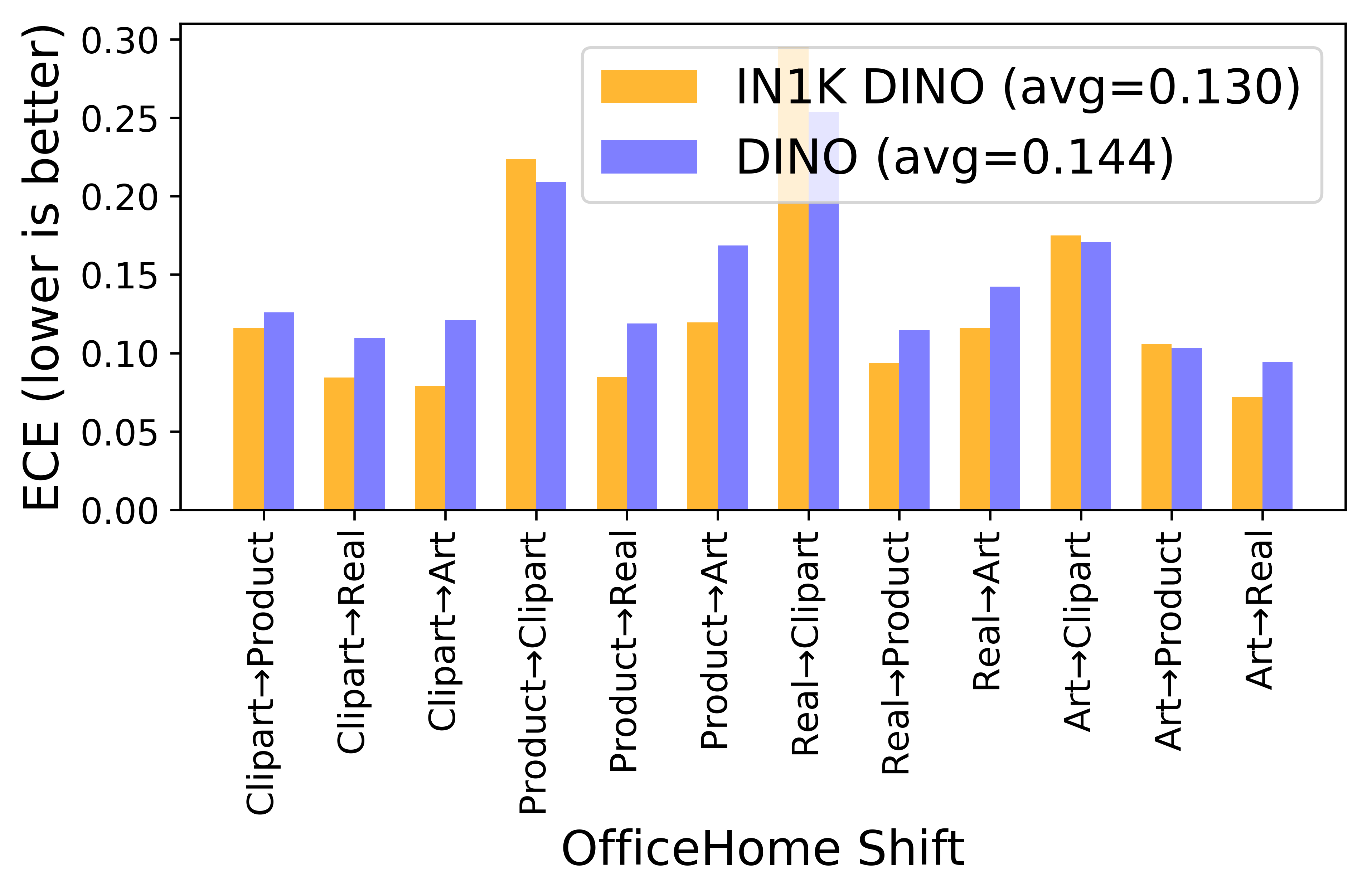}
            \caption[]%
            {{\small \textbf{DINO S+T pretraining}}}  
            \label{fig:dino_ood}
            \end{subfigure}
      \caption{\textbf{Effect of in-domain pretraining on OOD calibration}. Expected calibration on the target test set for each OfficeHome shifts (lower is better) }
      \label{fig:pretrain_ood}
    \vspace{-7pt}
  \end{figure}

In Fig.~\ref{fig:pretrain_ood} we analyze the effect of in-domain pretraining on out-of-distribution confidence calibration on the target test set after S+T pretraining with the MAE and DINO SSL strategies. We report expected calibration error (ECE~\cite{guo2017calibration}), lower is better. We observe inconsistent trends across shifts, with additional MAE pretraining improving out-of-distribution confidence calibration on 5/12 shifts on the OfficeHome benchmark, while DINO improving it only on 4/12.

\subsection{Encoder distance plots}

\begin{figure}[t]
    \centering  
      \begin{subfigure}[t]{0.48\textwidth}
        \centering
        \includegraphics[width=\linewidth]{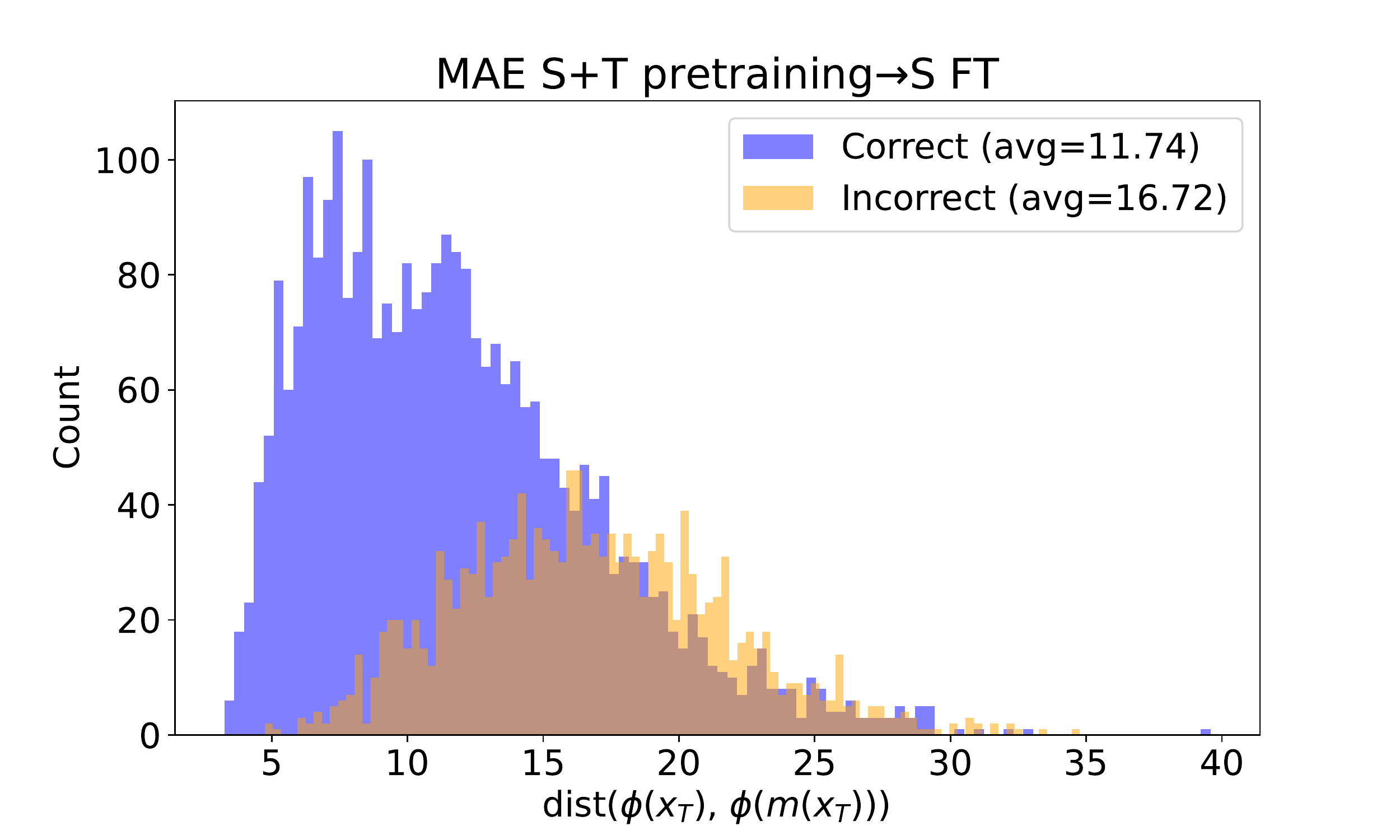}
        \caption[]%
        {{\small \textbf{MAE S+T pretraining}  }}  
        \label{fig:mae_dist}
        \end{subfigure}
        \begin{subfigure}[t]{0.48\textwidth}
            \centering
            \includegraphics[width=\linewidth]{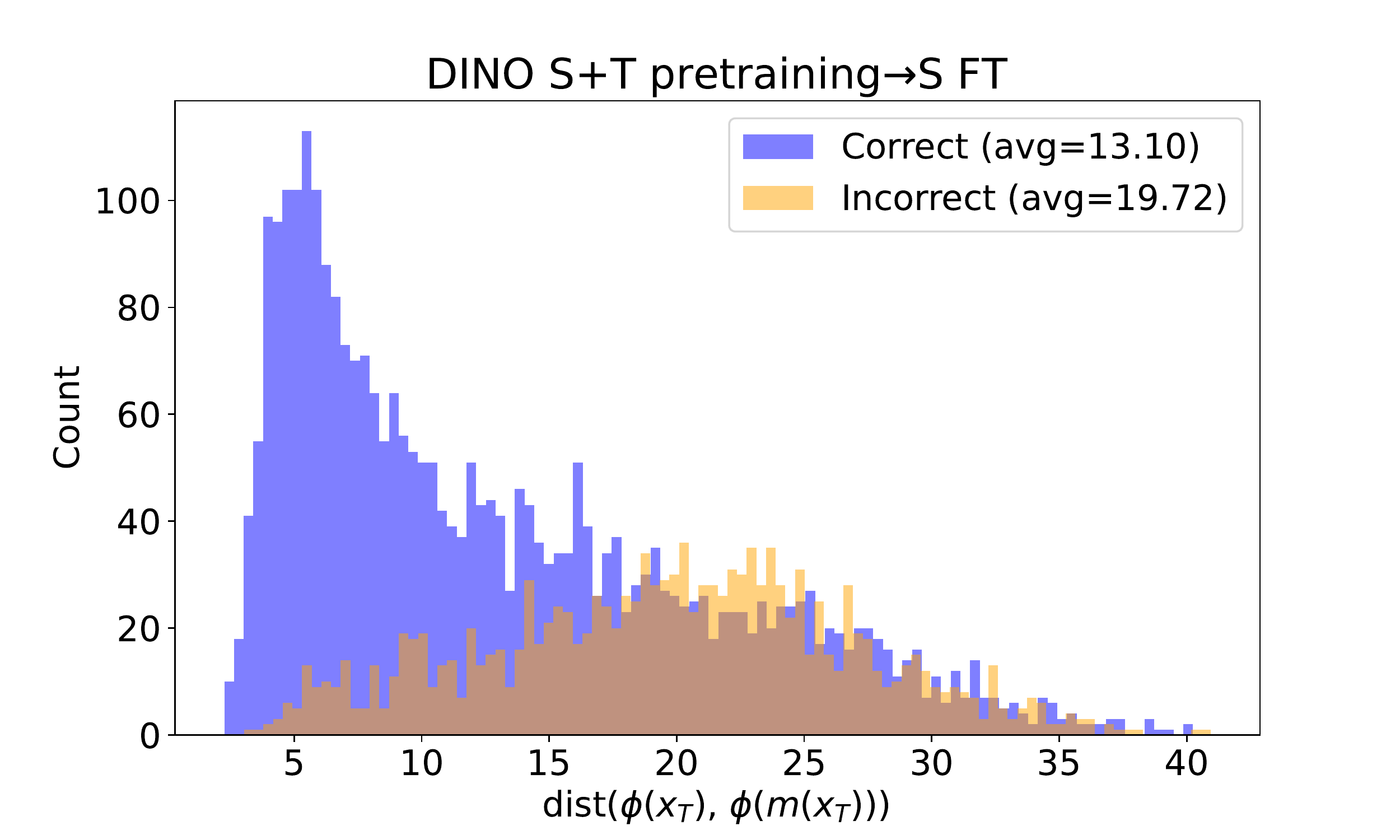}
            \caption[]%
            {{\small \textbf{DINO S+T pretraining}}}  
            \label{fig:dino_dist}
            \end{subfigure}
        \begin{subfigure}[t]{0.48\textwidth}
            \centering
            \includegraphics[width=\linewidth]{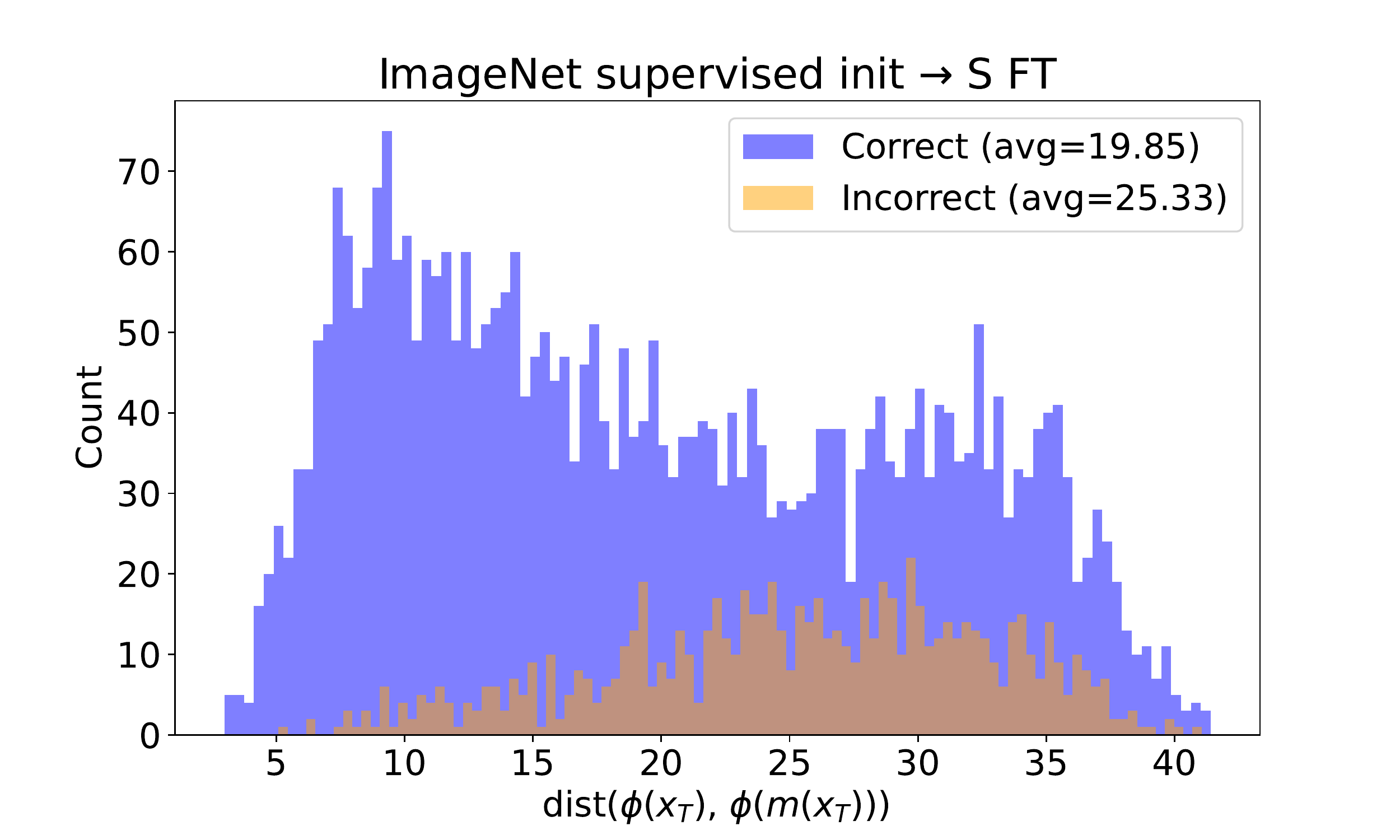}
            \caption[]%
            {{\small \textbf{Supervised ImageNet initialisation}}}  
            \label{fig:sup_dist}
            \end{subfigure}
      \caption{\textbf{Distribution of distance between the encoded representations of masked and original images}. If these distributions for correctly and incorrectly predicted target samples are well separated, target selection based on consistency is expected to work better. Numbers in legend denote average distance between embeddings for original and masked image.}
    \label{fig:pretrain_encoder_dist}
    \vspace{-7pt}
  \end{figure}

In Fig.~\ref{fig:pretrain_encoder_dist} we visualize histograms of the distance between class token embeddings extracted from the last transformer encoder layer, for target instances without and with random masks. We visualize these histograms for models finetuned on the source domains but with different initializations -- SSL initializations with additional S+T pretraining with MAE (Fig.~\ref{fig:mae_dist}) and DINO (Fig.~\ref{fig:dino_dist}), and a supervised ImageNet initialization. Instances that are classifier correctly and incorrectly are shown separately. As seen, with SSL initializations correct instances tend to on average have more similar embeddings across masking than supervised initializations (as a result of being trained to learn from such missing inputs during SSL pretraining), but this is not the case for the supervised initialization. This explains the efficacy of our masking consistency-based reliability scheme for SSL initializations.

\subsection{In-domain pretraining: t-SNE~\cite{van2008visualizing} visualization}

\begin{figure}
    \centering
    \includegraphics[width=\textwidth]{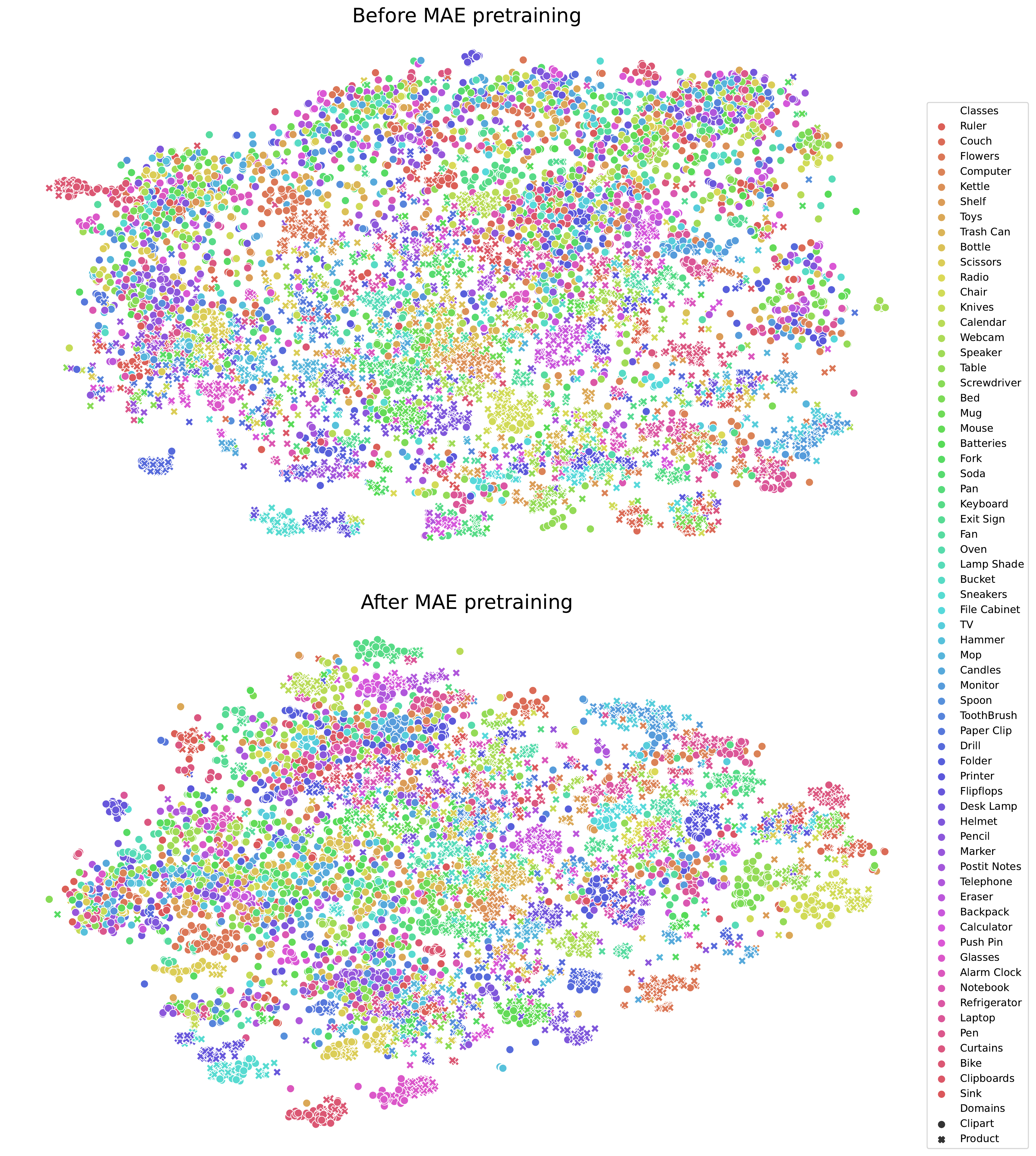}
    \caption{t-SNE visualization of CLS token features of images from Clipart and Product domains of OfficeHome before and after in-domain pretraining with MAE. }
    \label{fig:mae_tsne}
\end{figure}

\begin{figure}
    \centering
    \includegraphics[width=\textwidth]{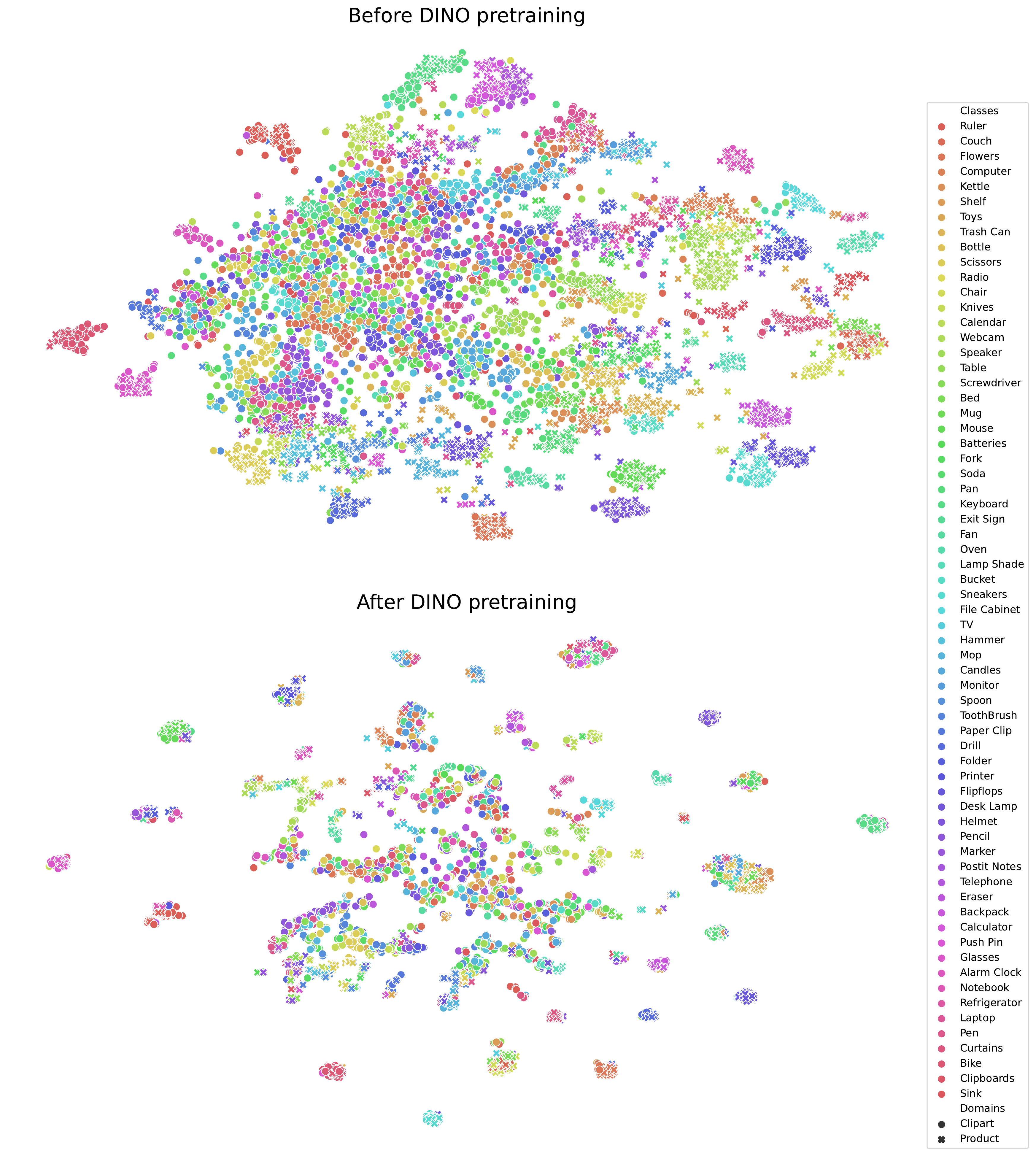}
    \caption{t-SNE visualization of CLS token features of images from Clipart and Product domains of OfficeHome before and after in-domain pretraining with DINO.}
    \label{fig:dino_tsne}
\end{figure}

In Figures~\ref{fig:mae_tsne}-~\ref{fig:dino_tsne}, we present t-SNE visualizations of class token activations from the encoder, for the Clipart and Product OfficeHome domains. We separately visualize features before and after in-domain pretraining with MAE~\ref{fig:mae_tsne} and DINO~\ref{fig:dino_tsne}. We note that these features are completely self-supervised as the model has not seen task labels yet. Regardless, we observe a small degree of task discriminativeness (examples of the same class are clustered together) and domain invariance (examples of the same class but different domains are close) before additional pretraining. After pretraining, we observe it to increase, particularly after DINO pretraining.

\subsection{Variance across runs}

To measure the performance variance of our method \method, we run it across 5 random seeds on the OfficeHome Clipart$\to$Product shift and observe a target accuracy mean of 73.1\% with a standard deviation of 0.59\%.

\section{Supervised ImageNet initializations}

\subsection{\method performance}

For completeness, we benchmark \method on top of a supervised ImageNet initialization, and find it to improve average accuracy across 12 OfficeHome shifts from 75.1\% to 76.8\%. However, we find a competing method designed for supervised initializations, TVT~\cite{yang2021tvt}, to obtain an average accuracy of 80.1\%, clearly outperforming our method, despite strongly underperforming \method with MAE and DINO initializations (Tables 1-3 in the main paper). This illustrates the importance of learning representations from missing information during pretraining, in the absence of which predictive consistency across masking proves to be an ineffective reliability measure. In Fig.~\ref{fig:pretrain_encoder_dist}(c) we further highlight this behavior.

\subsection{Most existing DA methods do not truly evaluate domain adaptation}

Most existing DA methods are initialized with supervised ImageNet initializations (with ViTs, mostly on ImageNet-22K), and adapted to standard benchmarks like OfficeHome, DomainNet, and VisDA. We now measure the degree of label overlap between ImageNet-22K and these 3 benchmarks. \textcolor{blue}{Astonishingly, the overlap is near 100\%: 61/65 (OfficeHome), 40/40 (DomainNet), and 12/12 (VisDA) categories from these benchmarks directly correspond to an ImageNet-22k category. This is particularly problematic when evaluating adaptation to real domains as a target; by definition DA assumes that the model has never seen labeled images from the target domain, but we argue that methods initialization with supervised ImageNet pretraining and adapted to real domains have seen plenty!} However, in this paper, all methods are initialized with self-supervised ImageNet initializations and we thus present a more realistic measure of adaptation performance, even when the target domain contains real images. Going forward, we urge the community to rethink DA benchmarking when starting with supervised initializations, and consider self-supervised initializations as a potentially fairer alternative starting point.

Self-supervised initializations may also be a superior initialization choice when adapting to domains very different from ImageNet. For \textit{e.g.,} Kim~\emph{et al.}~\cite{kim2022broad} find that SSL initializations outperform supervised ones when generalizing to a benchmark like WILDS~\cite{koh2021wilds}, which contains very distinct images from ImageNet. Similarly, Azizi~\emph{et al.}~\cite{azizi2021big} find that self-supervised pretraining on ImageNet followed by domain-specific pretraining strongly outperforms supervised ImageNet pretraining for medical image classification tasks.

\section{Datasets and implementation details}

\textbf{Data Licenses:} Images from the OfficeHome, DomainNet, and VisDA datasets are freely available for non-commercial and academic use. The creators note that while the datasets contain some copyrighted material, scientific research is considered a fair use of such material. To the best of our knowledge, none of the above datasets contain personally identifiable information or offensive content. 

\noindent\textbf{Hyperparameters.} Tables~\ref{tab:pt_hp}-~\ref{tab:impl_mae_finetune} include a detailed list of hyperparameters for the in-domain pretraining and adaptation phases.

\noindent\textbf{Compute details.} For most experiments, we use NVIDIA A40 GPUs (4 for pretraining and 1 for finetuning/adaptation) on an internal compute cluster.

\newcolumntype{x}[1]{>{\centering\arraybackslash}p{#1pt}}
\newcolumntype{y}[1]{>{\raggedright\arraybackslash}p{#1pt}}
\newcolumntype{z}[1]{>{\raggedleft\arraybackslash}p{#1pt}}
\newlength\savewidth\newcommand\shline{\noalign{\global\savewidth\arrayrulewidth
  \global\arrayrulewidth 1pt}\hline\noalign{\global\arrayrulewidth\savewidth}}

\begin{table}

\begin{subtable}{.48\textwidth}
\tablestyle{6pt}{1.02}
\scriptsize
\centering
\begin{tabular}{l | l}
config & value \\
\shline
optimizer & AdamW \cite{loshchilov2018decoupled} \\
initial learning rate & 4e-4 \\
final learning rate & 0 \\
weight decay & 0.05 \\
optimizer momentum & $\beta_1, \beta_2{=}0.9, 0.95$ \\
batch size & 2048 \\
learning rate schedule & cosine decay \cite{loshchilov2017sgdr} \\
warmup epochs \cite{goyal2017accurate} & 40 \\
augmentation & RandomResizedCrop \\
&+ RandAugment(3, 4) \cite{cubuk2020randaugment} \\
epochs & 800 (50 for VisDA) \\
drop path rate & 0.0 \\
\end{tabular}
\vspace{-.5em}
\caption{\textbf{MAE pretraining}}
\label{tab:impl_mae_pretrain} \vspace{-.5em}
\end{subtable}
\hfill
\begin{subtable}{.48\textwidth}
\tablestyle{6pt}{1.02}
\scriptsize
\centering
\begin{tabular}{l | l}
config & value \\
\shline
optimizer & AdamW \cite{loshchilov2018decoupled} \\
initial learning rate & 5e-5 \\
final learning rate & 3.8e-5 (1e-6 when OH Art is target) \\
initial weight decay & 0.04 \\
final weight decay & 0.16 (0.4 when OH Art is target) \\
optimizer momentum & $\beta_1, \beta_2{=}0.9, 0.999$  \\
batch size & 256 \\
learning rate schedule & cosine decay \cite{loshchilov2017sgdr} \\
warmup epochs \cite{goyal2017accurate} & 50 (10 when OH Art is target) \\
augmentation & RandomResizedCrop, ColorJitter \\
    & Solarization, GaussianBlur \\
epochs & 200 \\
drop path rate \cite{huang2016stochastic} & 0.1
\end{tabular}
\vspace{-.5em}
\caption{\textbf{DINO pretraining}}
\label{tab:impl_dino_pretrain} \vspace{-.5em}
\end{subtable}
\caption{Pretraining hyperparameters}
\label{tab:pt_hp}
\end{table}

\begin{table}[t]
\tablestyle{6pt}{1.02}
\scriptsize
\begin{tabular}{y{96}|y{90}}
config & value \\
\shline
optimizer & AdamW \cite{loshchilov2018decoupled} \\
initial learning rate & 2e-4 \\
weight decay & 0.05 \\
optimizer momentum & $\beta_1, \beta_2{=}0.9, 0.999$ \\
layer-wise lr decay \cite{clark2020electric,bao2021beit} & 0.75 \\
batch size & 1024 (for fine-tuning) and 512 (for adaptation) \\
learning rate schedule & constant \\
warmup epochs & 5 \\
training epochs & 100 (10 for VisDA) \\
augmentation & RandomResizedCrop \\
    & + RandAugment (1, 2.0)~\cite{cubuk2020randaugment} \\
label smoothing \cite{szegedy2016when} & 0.1 \\
drop path rate \cite{huang2016stochastic} & 0.1  \\
\end{tabular}
\vspace{-.5em}
\caption{\textbf{Fine-tuning/adaptation hyperparameters}}
\label{tab:impl_mae_finetune} \vspace{-.5em}
\end{table}

\section{DINO and MAE Self-supervised objectives}

Recall that in Sec. 3.2 we presented two types of SSL strategies for ViT's based on masked image modeling and joint embeddings. Both optimize an objective of the form $\mathcal{L}_{SSL}(\featext(m(\mathbf{x}_\target)), \featext(\mathbf{x}_\target))$, which could correspond to a reconstruction or invariance based objective operating on encoded features. We now instantiate $\mathcal{L}_{SSL}$ in the context of the the two SSL strategies we use in the paper: MAE~\cite{he2016deep} and DINO~\cite{caron2021emerging}.

\noindent\textbf{MAE~\cite{he2016deep}.} MAE learns a visual transformer autoencoder to reconstruct images $\mathbf{x}_\target$ given only a random subset of patches from the original image $m(\mathbf{x}_\target)$. Let $d$ denote a decoder learned for image reconstruction. The MAE objective minimizes the mean square error between the original and reconstructed image in pixel space:

\begin{equation}
    \mathcal{L}_{SSL}(\mathbf{x}_\target) = || \mathbf{x}_\target - d(\featext(m(\mathbf{x}_\target))) ||_2    
    \label{eq:mae}
\end{equation}

\noindent\textbf{DINO~\cite{caron2021emerging}.} For a target image $\mathbf{x}_\target$, DINO passes two transformed versions of each target image $m_1(\mathbf{x}_\target)$ and $m_2(\mathbf{x}_\target)$ to a student network $\featext_s$ and teacher network $\featext_t$ respectively, and trains the student network to predict the output of the teacher by minimizing the cross-entropy $H$ between the two output distributions:

\begin{equation}
    \mathcal{L}_{SSL}(\mathbf{x}_\target) = H(\featext_t(m_1(\mathbf{x}_\target)), \featext_s(m_2(\mathbf{x}_\target)))
    \label{eq:dino}
\end{equation}

%% file: neurips_2022.bbl
\begin{thebibliography}{10}

\bibitem{torralba2011unbiased}
A.~Torralba and A.~A. Efros, ``Unbiased look at dataset bias,'' in {\em CVPR
  2011}, pp.~1521--1528, IEEE, 2011.

\bibitem{saenko2010adapting}
K.~Saenko, B.~Kulis, M.~Fritz, and T.~Darrell, ``Adapting visual category
  models to new domains,'' in {\em European conference on computer vision},
  pp.~213--226, Springer, 2010.

\bibitem{tzeng2014deep}
E.~Tzeng, J.~Hoffman, N.~Zhang, K.~Saenko, and T.~Darrell, ``Deep domain
  confusion: Maximizing for domain invariance,'' {\em arXiv preprint
  arXiv:1412.3474}, 2014.

\bibitem{long2015learning}
M.~Long, Y.~Cao, J.~Wang, and M.~Jordan, ``Learning transferable features with
  deep adaptation networks,'' in {\em International conference on machine
  learning}, pp.~97--105, PMLR, 2015.

\bibitem{tzeng2017adversarial}
E.~Tzeng, J.~Hoffman, K.~Saenko, and T.~Darrell, ``Adversarial discriminative
  domain adaptation,'' in {\em Proceedings of the IEEE conference on computer
  vision and pattern recognition}, pp.~7167--7176, 2017.

\bibitem{ganin2015unsupervised}
Y.~Ganin and V.~Lempitsky, ``Unsupervised domain adaptation by
  backpropagation,'' in {\em International conference on machine learning},
  pp.~1180--1189, PMLR, 2015.

\bibitem{long2018conditional}
M.~Long, Z.~Cao, J.~Wang, and M.~I. Jordan, ``Conditional adversarial domain
  adaptation,'' {\em Advances in neural information processing systems},
  vol.~31, 2018.

\bibitem{hoffman2018cycada}
J.~Hoffman, E.~Tzeng, T.~Park, J.-Y. Zhu, P.~Isola, K.~Saenko, A.~Efros, and
  T.~Darrell, ``Cycada: Cycle-consistent adversarial domain adaptation,'' in
  {\em International conference on machine learning}, pp.~1989--1998, PMLR,
  2018.

\bibitem{prabhu2021sentry}
V.~Prabhu, S.~Khare, D.~Kartik, and J.~Hoffman, ``Sentry: Selective entropy
  optimization via committee consistency for unsupervised domain adaptation,''
  in {\em Proceedings of the IEEE/CVF International Conference on Computer
  Vision}, pp.~8558--8567, 2021.

\bibitem{russakovsky2015imagenet}
O.~Russakovsky, J.~Deng, H.~Su, J.~Krause, S.~Satheesh, S.~Ma, Z.~Huang,
  A.~Karpathy, A.~Khosla, M.~Bernstein, {\em et~al.}, ``Imagenet large scale
  visual recognition challenge,'' {\em International journal of computer
  vision}, vol.~115, no.~3, pp.~211--252, 2015.

\bibitem{dosovitskiy2020image}
A.~Dosovitskiy, L.~Beyer, A.~Kolesnikov, D.~Weissenborn, X.~Zhai,
  T.~Unterthiner, M.~Dehghani, M.~Minderer, G.~Heigold, S.~Gelly, {\em et~al.},
  ``An image is worth 16x16 words: Transformers for image recognition at
  scale,'' in {\em International Conference on Learning Representations}, 2020.

\bibitem{khan2021transformers}
S.~Khan, M.~Naseer, M.~Hayat, S.~W. Zamir, F.~S. Khan, and M.~Shah,
  ``Transformers in vision: A survey,'' {\em ACM Computing Surveys (CSUR)},
  2021.

\bibitem{minderer2021revisiting}
M.~Minderer, J.~Djolonga, R.~Romijnders, F.~Hubis, X.~Zhai, N.~Houlsby,
  D.~Tran, and M.~Lucic, ``Revisiting the calibration of modern neural
  networks,'' {\em Advances in Neural Information Processing Systems}, vol.~34,
  2021.

\bibitem{yang2021tvt}
J.~Yang, J.~Liu, N.~Xu, and J.~Huang, ``Tvt: Transferable vision transformer
  for unsupervised domain adaptation,'' {\em arXiv preprint arXiv:2108.05988},
  2021.

\bibitem{xu2021cdtrans}
T.~Xu, W.~Chen, P.~Wang, F.~Wang, H.~Li, and R.~Jin, ``Cdtrans: Cross-domain
  transformer for unsupervised domain adaptation,'' {\em arXiv preprint
  arXiv:2109.06165}, 2021.

\bibitem{raghu2019transfusion}
M.~Raghu, C.~Zhang, J.~Kleinberg, and S.~Bengio, ``Transfusion: Understanding
  transfer learning for medical imaging,'' {\em Advances in neural information
  processing systems}, vol.~32, 2019.

\bibitem{azizi2021big}
S.~Azizi, B.~Mustafa, F.~Ryan, Z.~Beaver, J.~Freyberg, J.~Deaton, A.~Loh,
  A.~Karthikesalingam, S.~Kornblith, T.~Chen, {\em et~al.}, ``Big
  self-supervised models advance medical image classification,'' in {\em
  Proceedings of the IEEE/CVF International Conference on Computer Vision},
  pp.~3478--3488, 2021.

\bibitem{doersch2015unsupervised}
C.~Doersch, A.~Gupta, and A.~A. Efros, ``Unsupervised visual representation
  learning by context prediction,'' in {\em Proceedings of the IEEE
  international conference on computer vision}, pp.~1422--1430, 2015.

\bibitem{wang2015unsupervised}
X.~Wang and A.~Gupta, ``Unsupervised learning of visual representations using
  videos,'' in {\em Proceedings of the IEEE international conference on
  computer vision}, pp.~2794--2802, 2015.

\bibitem{noroozi2016unsupervised}
M.~Noroozi and P.~Favaro, ``Unsupervised learning of visual representations by
  solving jigsaw puzzles,'' in {\em European conference on computer vision},
  pp.~69--84, Springer, 2016.

\bibitem{wu2018unsupervised}
Z.~Wu, Y.~Xiong, S.~X. Yu, and D.~Lin, ``Unsupervised feature learning via
  non-parametric instance discrimination,'' in {\em Proceedings of the IEEE
  conference on computer vision and pattern recognition}, pp.~3733--3742, 2018.

\bibitem{chen2020simple}
T.~Chen, S.~Kornblith, M.~Norouzi, and G.~Hinton, ``A simple framework for
  contrastive learning of visual representations,'' in {\em International
  conference on machine learning}, pp.~1597--1607, PMLR, 2020.

\bibitem{he2020momentum}
K.~He, H.~Fan, Y.~Wu, S.~Xie, and R.~Girshick, ``Momentum contrast for
  unsupervised visual representation learning,'' in {\em Proceedings of the
  IEEE/CVF conference on computer vision and pattern recognition},
  pp.~9729--9738, 2020.

\bibitem{caron2020unsupervised}
M.~Caron, I.~Misra, J.~Mairal, P.~Goyal, P.~Bojanowski, and A.~Joulin,
  ``Unsupervised learning of visual features by contrasting cluster
  assignments,'' {\em Advances in Neural Information Processing Systems},
  vol.~33, pp.~9912--9924, 2020.

\bibitem{caron2021emerging}
M.~Caron, H.~Touvron, I.~Misra, H.~J{\'e}gou, J.~Mairal, P.~Bojanowski, and
  A.~Joulin, ``Emerging properties in self-supervised vision transformers,'' in
  {\em Proceedings of the IEEE/CVF International Conference on Computer
  Vision}, pp.~9650--9660, 2021.

\bibitem{he2021masked}
K.~He, X.~Chen, S.~Xie, Y.~Li, P.~Doll{\'a}r, and R.~Girshick, ``Masked
  autoencoders are scalable vision learners,'' {\em arXiv preprint
  arXiv:2111.06377}, 2021.

\bibitem{kim2021cds}
D.~Kim, K.~Saito, T.-H. Oh, B.~A. Plummer, S.~Sclaroff, and K.~Saenko, ``Cds:
  Cross-domain self-supervised pre-training,'' in {\em Proceedings of the
  IEEE/CVF International Conference on Computer Vision}, pp.~9123--9132, 2021.

\bibitem{shen2022connect}
K.~Shen, R.~Jones, A.~Kumar, S.~M. Xie, J.~Z. HaoChen, T.~Ma, and P.~Liang,
  ``Connect, not collapse: Explaining contrastive learning for unsupervised
  domain adaptation,'' {\em arXiv preprint arXiv:2204.00570}, 2022.

\bibitem{assran2022masked}
M.~Assran, M.~Caron, I.~Misra, P.~Bojanowski, F.~Bordes, P.~Vincent, A.~Joulin,
  M.~Rabbat, and N.~Ballas, ``Masked siamese networks for label-efficient
  learning,'' {\em arXiv preprint arXiv:2204.07141}, 2022.

\bibitem{venkateswara2017deep}
H.~Venkateswara, J.~Eusebio, S.~Chakraborty, and S.~Panchanathan, ``Deep
  hashing network for unsupervised domain adaptation,'' in {\em Proceedings of
  the IEEE conference on computer vision and pattern recognition},
  pp.~5018--5027, 2017.

\bibitem{peng2019moment}
X.~Peng, Q.~Bai, X.~Xia, Z.~Huang, K.~Saenko, and B.~Wang, ``Moment matching
  for multi-source domain adaptation,'' in {\em Proceedings of the IEEE/CVF
  international conference on computer vision}, pp.~1406--1415, 2019.

\bibitem{peng2017visda}
X.~Peng, B.~Usman, N.~Kaushik, J.~Hoffman, D.~Wang, and K.~Saenko, ``Visda: The
  visual domain adaptation challenge,'' {\em arXiv preprint arXiv:1710.06924},
  2017.

\bibitem{tan2020class}
S.~Tan, X.~Peng, and K.~Saenko, ``Class-imbalanced domain adaptation: an
  empirical odyssey,'' in {\em European Conference on Computer Vision},
  pp.~585--602, Springer, 2020.

\bibitem{bao2021beit}
H.~Bao, L.~Dong, and F.~Wei, ``Beit: Bert pre-training of image transformers,''
  {\em arXiv preprint arXiv:2106.08254}, 2021.

\bibitem{pathak2016context}
D.~Pathak, P.~Krahenbuhl, J.~Donahue, T.~Darrell, and A.~A. Efros, ``Context
  encoders: Feature learning by inpainting,'' in {\em Proceedings of the IEEE
  conference on computer vision and pattern recognition}, pp.~2536--2544, 2016.

\bibitem{he2016deep}
K.~He, X.~Zhang, S.~Ren, and J.~Sun, ``Deep residual learning for image
  recognition,'' in {\em Proceedings of the IEEE conference on computer vision
  and pattern recognition}, pp.~770--778, 2016.

\bibitem{xie2021simmim}
Z.~Xie, Z.~Zhang, Y.~Cao, Y.~Lin, J.~Bao, Z.~Yao, Q.~Dai, and H.~Hu, ``Simmim:
  A simple framework for masked image modeling,'' {\em arXiv preprint
  arXiv:2111.09886}, 2021.

\bibitem{wei2021masked}
C.~Wei, H.~Fan, S.~Xie, C.-Y. Wu, A.~Yuille, and C.~Feichtenhofer, ``Masked
  feature prediction for self-supervised visual pre-training,'' {\em arXiv
  preprint arXiv:2112.09133}, 2021.

\bibitem{grandvalet2004semi}
Y.~Grandvalet and Y.~Bengio, ``Semi-supervised learning by entropy
  minimization,'' {\em Advances in neural information processing systems},
  vol.~17, 2004.

\bibitem{chen2019progressive}
C.~Chen, W.~Xie, W.~Huang, Y.~Rong, X.~Ding, Y.~Huang, T.~Xu, and J.~Huang,
  ``Progressive feature alignment for unsupervised domain adaptation,'' in {\em
  Proceedings of the IEEE/CVF Conference on Computer Vision and Pattern
  Recognition}, pp.~627--636, 2019.

\bibitem{jiang2020implicit}
X.~Jiang, Q.~Lao, S.~Matwin, and M.~Havaei, ``Implicit class-conditioned domain
  alignment for unsupervised domain adaptation,'' in {\em International
  Conference on Machine Learning}, pp.~4816--4827, PMLR, 2020.

\bibitem{zou2018unsupervised}
Y.~Zou, Z.~Yu, B.~Kumar, and J.~Wang, ``Unsupervised domain adaptation for
  semantic segmentation via class-balanced self-training,'' in {\em Proceedings
  of the European conference on computer vision (ECCV)}, pp.~289--305, 2018.

\bibitem{kim2022ev}
J.~Kim, I.~Hwang, and Y.~M. Kim, ``Ev-tta: Test-time adaptation for event-based
  object recognition,'' {\em arXiv preprint arXiv:2203.12247}, 2022.

\bibitem{prabhu2021s4t}
V.~Prabhu, S.~Khare, D.~Kartik, and J.~Hoffman, ``S4t: Source-free domain
  adaptation for semantic segmentation via self-supervised selective
  self-training,'' {\em arXiv preprint arXiv:2107.10140}, 2021.

\bibitem{loshchilov2018decoupled}
I.~Loshchilov and F.~Hutter, ``Decoupled weight decay regularization,'' in {\em
  International Conference on Learning Representations}, 2018.

\bibitem{cubuk2020randaugment}
E.~D. Cubuk, B.~Zoph, J.~Shlens, and Q.~V. Le, ``Randaugment: Practical
  automated data augmentation with a reduced search space,'' in {\em
  Proceedings of the IEEE/CVF Conference on Computer Vision and Pattern
  Recognition Workshops}, pp.~702--703, 2020.

\bibitem{paszke2019pytorch}
A.~Paszke, S.~Gross, F.~Massa, A.~Lerer, J.~Bradbury, G.~Chanan, T.~Killeen,
  Z.~Lin, N.~Gimelshein, L.~Antiga, {\em et~al.}, ``Pytorch: An imperative
  style, high-performance deep learning library,'' {\em Advances in neural
  information processing systems}, vol.~32, 2019.

\bibitem{jin2020minimum}
Y.~Jin, X.~Wang, M.~Long, and J.~Wang, ``Minimum class confusion for versatile
  domain adaptation,'' in {\em European Conference on Computer Vision},
  pp.~464--480, Springer, 2020.

\bibitem{kim2022broad}
D.~Kim, K.~Wang, S.~Sclaroff, and K.~Saenko, ``A broad study of pre-training
  for domain generalization and adaptation,'' {\em arXiv preprint
  arXiv:2203.11819}, 2022.

\bibitem{van2008visualizing}
L.~Van~der Maaten and G.~Hinton, ``Visualizing data using t-sne.,'' {\em
  Journal of machine learning research}, vol.~9, no.~11, 2008.

\bibitem{guo2017calibration}
C.~Guo, G.~Pleiss, Y.~Sun, and K.~Q. Weinberger, ``On calibration of modern
  neural networks,'' in {\em International Conference on Machine Learning},
  pp.~1321--1330, PMLR, 2017.

\bibitem{koh2021wilds}
P.~W. Koh, S.~Sagawa, H.~Marklund, S.~M. Xie, M.~Zhang, A.~Balsubramani, W.~Hu,
  M.~Yasunaga, R.~L. Phillips, I.~Gao, {\em et~al.}, ``Wilds: A benchmark of
  in-the-wild distribution shifts,'' in {\em International Conference on
  Machine Learning}, pp.~5637--5664, PMLR, 2021.

\bibitem{loshchilov2017sgdr}
I.~Loshchilov and F.~Hutter, ``{SGDR:} stochastic gradient descent with warm
  restarts,'' in {\em 5th International Conference on Learning Representations,
  {ICLR} 2017, Toulon, France, April 24-26, 2017, Conference Track
  Proceedings}, OpenReview.net, 2017.

\bibitem{goyal2017accurate}
P.~Goyal, P.~Dollár, R.~Girshick, P.~Noordhuis, L.~Wesolowski, A.~Kyrola,
  A.~Tulloch, Y.~Jia, and K.~He, ``Accurate, large minibatch sgd: Training
  imagenet in 1 hour,'' 2017.

\bibitem{huang2016stochastic}
G.~Huang, Y.~Sun, Z.~Liu, D.~Sedra, and K.~Q. Weinberger, ``Deep networks with
  stochastic depth,'' in {\em Computer Vision -- ECCV 2016} (B.~Leibe,
  J.~Matas, N.~Sebe, and M.~Welling, eds.), (Cham), pp.~646--661, Springer
  International Publishing, 2016.

\bibitem{clark2020electric}
K.~Clark, M.-T. Luong, Q.~V. Le, and C.~D. Manning, ``Pre-training transformers
  as energy-based cloze models,'' in {\em EMNLP}, 2020.

\bibitem{szegedy2016when}
C.~Szegedy, V.~Vanhoucke, S.~Ioffe, J.~Shlens, and Z.~Wojna, ``Rethinking the
  inception architecture for computer vision,'' in {\em 2016 IEEE Conference on
  Computer Vision and Pattern Recognition (CVPR)}, pp.~2818--2826, 2016.

\end{thebibliography}
